\documentclass[sigconf]{acmart}
\AtBeginDocument{%
  }

\setcopyright{acmlicensed}
\copyrightyear{2018}
\acmYear{2018}
\acmDOI{XXXXXXX.XXXXXXX}
\acmConference[Conference acronym 'XX]{Make sure to enter the correct
  conference title from your rights confirmation email}{June 03--05,
  2018}{Woodstock, NY}
\acmISBN{978-1-4503-XXXX-X/2018/06}

\usepackage{graphicx}
\usepackage{color}
\usepackage{xcolor}
\usepackage{paralist}
\usepackage{xpatch}
\usepackage{multirow}
\usepackage[normalem]{ulem}
\usepackage[inline]{enumitem}
\usepackage{colortbl}
\usepackage{wrapfig,lipsum,booktabs}
\usepackage{subcaption}
\usepackage[font=small,skip=5pt]{caption}
\setlist[itemize]{noitemsep, topsep=0pt}

\definecolor{dashboxgreen}{rgb}{0.1059, 0.6588, 0.0039}
\definecolor{LightCyan}{rgb}{0.88,1,1}

\xpatchcmd{\sout}
  {\bgroup}
  {\bgroup}
  {}{}

\begin{document}
% \sloppy

%%
%% The "title" command has an optional parameter,
%% allowing the author to define a "short title" to be used in page headers.

\title{FedNoisy: Federated Noisy Label Learning Benchmark}

%%
%% The "author" command and its associated commands are used to define
%% the authors and their affiliations.
%% Of note is the shared affiliation of the first two authors, and the
%% "authornote" and "authornotemark" commands
%% used to denote shared contribution to the research.
\author{Siqi Liang}
\email{siqilian@umich.edu}
\affiliation{
  \institution{University of Michigan}
  \city{Ann Arbor}
  \state{Michigan}
  \country{USA}
}

\author{Jintao Huang}
\email{a764695611@gmail.com}
\affiliation{
  \institution{Artifical Intelligence Innovation and Incubation Institute, Fudan University}
  \city{Shanghai}
  \country{China}
}

\author{Junyuan Hong}
\email{hongju12@msu.edu}
\affiliation{
 \institution{Michigan State University}
 \city{East Lansing}
 \state{Michigan}
 \country{USA}}

\author{Dun Zeng}
\email{zengdun.cs@gmail.com}
\affiliation{
  \institution{Artifical Intelligence Innovation and Incubation Institute, Fudan University}
  \city{Shanghai}
  \country{China}}

\author{Jiayu Zhou}
% \authornotemark[1]
\authornote{Corresponding authors.}
\email{jiayuz@umich.edu}
\affiliation{
  \institution{University of Michigan}
  \city{Ann Arbor}
  \state{Michigan}
  \country{USA}
}

\author{Zenglin Xu}
\authornotemark[1]
\email{zenglinxu@fudan.edu.cn}
\affiliation{
  \institution{Artifical Intelligence Innovation and Incubation Institute, Fudan University}
  \institution{Shanghai Academy of Al for Science}
  \city{Shanghai}
  \country{China}}

% \author{Julius P. Kumquat}
% \affiliation{%
%   \institution{The Kumquat Consortium}
%   \city{New York}
%   \country{USA}}
% \email{jpkumquat@consortium.net}

%%
%% By default, the full list of authors will be used in the page
%% headers. Often, this list is too long, and will overlap
%% other information printed in the page headers. This command allows
%% the author to define a more concise list
%% of authors' names for this purpose.
\renewcommand{\shortauthors}{Liang et al.}

%%
%% The abstract is a short summary of the work to be presented in the
%% article.
\begin{abstract}
 %  Federated learning (FL) is a promising learning paradigm that learns from distributed clients without compromising data confidentiality.
  % The distributed and isolated nature of data in FL may be complicated by data quality, making it more vulnerable to noisy labels.
  % Many efforts exist to defend against the negative impacts of noisy labels in centralized or federated settings. 
  % However, there is a lack of a benchmark that comprehensively considers the impact of noisy labels in a wide variety of typical FL settings.
  % This work provides the first standardized benchmark that offers insights into representative federated noisy settings. 
  % Using this benchmark, we conduct comprehensive experiments to explore the characteristics of these data settings, which shed light on future methodology development of FL under noisy labels. 
  % We highlight the 20 typical settings for five datasets proposed in our benchmark and standardized simulation pipeline for federated noisy label learning. 
  % We expect this benchmark to greatly improve the capability of idea verification in federated learning with noisy labels.
  
Federated learning has gained popularity for distributed learning without aggregating sensitive data from clients.
But meanwhile, the distributed and isolated nature of data isolation may be complicated by data quality, making it more vulnerable to noisy labels.
Many efforts exist to defend against the negative impacts of noisy labels in centralized or federated settings. 
However, there is a lack of a benchmark that comprehensively considers the impact of noisy labels in a wide variety of typical FL settings.
In this work, we serve the first standardized benchmark that can help researchers fully explore potential federated noisy settings. 
Also, we conduct comprehensive experiments to explore the characteristics of these data settings and the comparison across baselines, which may guide method development in the future. 
We highlight the 20 basic settings for 6 datasets proposed in our benchmark and standardized simulation pipeline for federated noisy label learning, including implementations of 9 baselines. We hope this benchmark can facilitate idea verification in federated learning with noisy labels. \texttt{FedNoisy} is available at~\url{https://github.com/SMILELab-FL/FedNoisy}.
\end{abstract}

%%
%% The code below is generated by the tool at http://dl.acm.org/ccs.cfm.
%% Please copy and paste the code instead of the example below.
%%


%%
%% Keywords. The author(s) should pick words that accurately describe
%% the work being presented. Separate the keywords with commas.
%% A "teaser" image appears between the author and affiliation
%% information and the body of the document, and typically spans the
%% page.
% \begin{teaserfigure}
%   \includegraphics[width=\textwidth]{sampleteaser}
%   \caption{Seattle Mariners at Spring Training, 2010.}
%   \Description{Enjoying the baseball game from the third-base
%   seats. Ichiro Suzuki preparing to bat.}
%   \label{fig:teaser}
% \end{teaserfigure}

% \received{20 February 2007}
% \received[revised]{12 March 2009}
% \received[accepted]{5 June 2009}

%%
%% This command processes the author and affiliation and title
%% information and builds the first part of the formatted document.

\maketitle

\section{Introduction}

Federated learning (FL) is a flourishing machine learning research area in recent years, which enables collaborative learning among different data centers (e.g., medical institutes and financial organizations) or small data collectors (e.g., mobile phones and smart wearable devices) to achieve promising model performance under restrictions on data sharing. 
Given the benefits of privacy protection and the efficiency of knowledge sharing, FL has attracted considerable attention and has been applied in various application fields, such as image classification \cite{DBLP:conf/iclr/JeongYYH21, liunderstanding, DBLP:conf/bigdataconf/ZhangYYYGRM21, zhou2022fedfa}, FinTech \cite{zhang2023privacy, DBLP:series/lncs/LongT0Z20}, text classification \cite{DBLP:conf/emnlp/ZhangHQWX22, DBLP:journals/corr/abs-2212-10025}, graph classification \cite{he2021fedgraphnn}, human activity recognition \cite{8672262, DBLP:conf/mobisys/OuyangXZHX21}, and healthcare \cite{DBLP:conf/cvpr/Liu00DH21, DBLP:conf/nips/TerrailACGHLMMM22, DBLP:journals/jhir/XuGSWBW21}.

A common yet implicit assumption in most existing FL research is that the local dataset on each client is carefully annotated with high-quality labels. 
However, in many real-world settings, well-labeled datasets are expensive, time-consuming, and can be prohibitive to acquire \cite{9729424}, for example, in medical applications \cite{KARIMI2020101759}. 
% And more commonly, only unreliable local samples annotated by users can be collected on smart devices.
Also, there may exist malicious attackers in FL systems that use label-flipping attacks \cite{DBLP:series/lncs/Lyu00Y20} to intentionally manipulate local samples. 
In applications with low data quality, omitting clients with unreliable samples may significantly reduce training data, while manually cleaning data can induce prohibitive costs. 

% \textcolor{red}{the logic from last para to the next is not consistent.}
It is well-studied that noisy labels often lead to poor model accuracy and degraded generalization capability \cite{DBLP:conf/iclr/ZhangBHRV17}. 
% The problem can be further complicated with non-IID data distribution in the FL setting. 
The non-IID distribution in FL settings, as a result of intra-client class imbalance and inter-client heterogeneous distribution, may further complicate the noisy label and aggravate the negative impacts of noisy labels in learning. 
This complication makes existing centralized noisy label learning (CNLL) less effective and requires joint consideration of noisy label learning (NLL) and heterogeneity in FL.  
% Both two require to be considered with noisy label learning (NLL) simultaneously in FNLL, which hinders the direct deployment of existing centralized noisy label learning (CNLL) methods into FL.

Therefore, federated noisy label learning (FNLL) emerges to fill the urgent need to utilize data with noisy labels in the FL training process.
There are many recent efforts towards FNLL. However, they are evaluated under different experimental settings and are thus hard to comprehensively compare and choose in practical applications, due to the lack of standard benchmarks and pipelines. 
% We especially want to highlight the undesired variation of experiment designs that makes it hard to tell when an algorithm works well.
The differences in experimental settings are due to the following factors. 
% Also, there is a gap between CNLL and  FNLL in both baselines and noise settings. 
\textbf{1) Datasets.} Existing works on FNLL are evaluated on different datasets. For synthetic noise, \cite{DBLP:journals/expert/YangPBK22, DBLP:conf/cvpr/XuCQC22, DBLP:journals/corr/abs-2205-10110} evaluate on CIFAR-10 which is common in both FL and CNLL; \cite{DBLP:journals/corr/abs-2001-11359} uses PD-Tremor dataset and \cite{DBLP:journals/fgcs/DuanLCJH22} uses Adult \cite{kohavi1996scaling}, and % Credit \cite{dal2015calibrating}, 
HFP \cite{chicco2020machine} datasets, which are rarely used in CNLL \cite{9729424}. 
% Others may use FEMNIST, which is common in FL but not in CNLL. 
\textbf{2) Data partition schemes.} Though most works \cite{DBLP:journals/corr/abs-2106-13239, DBLP:conf/cikm/Tahmasebian0022, DBLP:conf/kdd/LiPH22} use common IID partition schemes, they have differences in non-IID treatments and therefore different data partitions. \cite{DBLP:journals/corr/abs-2205-10110} used Dirichlet partition, which is commonly used among FL works. \cite{DBLP:journals/corr/abs-2001-11359} used real-world partition generated by hospitals. \cite{DBLP:conf/icpr/Tuor0KLL20} deployed an extreme distribution scheme that allows only a single class on each client. 
\textbf{3) Types of label noise.} Existing FNLL works were studied under a variety of generation processes of label noise in the FL environment. 
\cite{DBLP:journals/expert/YangPBK22, DBLP:journals/corr/abs-2205-10110} evaluated both symmetric noise and asymmetric label noise as in centralized NLL research. \cite{DBLP:conf/cvpr/XuCQC22} only evaluated symmetric noise; 
\cite{DBLP:conf/icpr/Tuor0KLL20} evaluated open-set noise and \cite{DBLP:series/lncs/0013RZJS20} used other types of noise. 
When it comes to noise distribution on different clients, different works introduced different settings: \cite{DBLP:journals/corr/abs-2001-11359, DBLP:conf/icpr/Tuor0KLL20, DBLP:conf/cikm/Tahmasebian0022} allowed clients to be either clean or noised, with all noised clients contain the same level of label noise; \cite{DBLP:journals/corr/abs-2205-10110} separated clients into 4 groups with 4 different noise levels; \cite{DBLP:journals/fgcs/DuanLCJH22} set all clients with same noise ratio.
\textbf{4) Limited baselines.} Most existing FNLL works combine only SOTA CNLL methods with FL algorithms as baselines in experiments, partially due to the fact that the implementations of FNLL approaches did not conform to a standard benchmark/pipeline, which makes consistent comparisons across FNLL hardly possible. 
For example, \cite{DBLP:conf/cvpr/XuCQC22} combined DivideMix\cite{DBLP:conf/iclr/LiSH20}/JointOpt\cite{DBLP:conf/cvpr/TanakaIYA18} with FedAvg as baselines; \cite{DBLP:journals/expert/YangPBK22} deployed Co-teaching on FedAvg as baseline. Most of these methods belong to limited categories in CNLL, i.e., sample selection methods \cite{9729424}, while dismissing others in CNLL. 
These limited baselines could lead to barriers when diving deep into FNLL, without fully exploring the potential capability of existing NLL methods. 
We attribute the scarcity of baselines to the lack of a well-designed FL framework, which includes a variety of datasets, easy access to different partition schemes, ready-to-go simulation environments, and extensible baselines covering a wide scope of methodologies.
% Also, different algorithm settings and implementations increase the challenges in FNLL research.

% This work aims to close the gap between existing CNLL research paradigms, including widely accepted datasets as well as representative methods from different lines, and the FL community that has already reached a consensus on federated scenes with clean data while not yet with label noise. Here we present \texttt{FedNoisy}, the first comprehensive FNLL benchmark, and our contributions can be concluded as follow:
% \begin{itemize}
%     \item We propose a standardized FNLL data setting generation pipeline, supporting at least 20 different federated scenes with different partition manners and generation mechanisms of label noise. And currently, we have included 4 datasets with built-in interfaces in \texttt{FedNoisy}. Also, we provide compatible pipeline interfaces and flexible functions that enable other public datasets to generate FNLL data settings with minimum effort.   
%     \item We implement an open-source FNLL codebase that comprises more than 4 FNLL baselines covering a broad scope of existing CNLL research categories. All baselines are presented in a unified FL workflow, capable of further improvement with different FL algorithms in a pluggable manner.
%     \item We conduct comprehensive evaluation following standardized experimental pipeline to show the characteristics of our FNLL data settings and effectiveness of our baselines.
% \end{itemize}
This work aims to fill the gap in existing FNLL research paradigms with a standardized setting on widely accepted datasets. 
% and \comm{??: the FL community that has already reached a consensus on federated scenes with clean data while not yet with label noise}. 
% Thus, we propose the first comprehensive FNLL benchmark, \texttt{FedNoisy}. 
Our contributions can be concluded as follow:
\begin{itemize}
    \item We propose a standardized FNLL benchmark, \texttt{FedNoisy}, including a comprehensive pipeline to generate noisy-label and heterogeneous FL settings.
    In total, the benchmark supports 20 different federated scenes covering a broad scope of different data partition manners and generation mechanisms of label noise. 
    % And currently, we have included 4 datasets with built-in interfaces in \texttt{FedNoisy}. 
    % Also, we provide compatible pipeline interfaces and flexible functions that enable other public datasets to generate FNLL data settings with minimum effort.
    \item We implement an FNLL codebase with ready-to-go federated noisy-label settings. The codebase comprises 9 FNLL baselines covering a broad scope of existing CNLL research categories as well as the most recent FNLL work. 
    All baselines are presented in a unified FL workflow, capable of further improvement with different FL algorithms in a pluggable manner. 
    We have included 6 datasets with built-in interfaces for the convenience of evaluating existing methods. 
    \item We conduct a comprehensive evaluation following a standardized experimental pipeline to show the intriguing characteristics of FNLL data settings as well as the comparison between the included FNLL baselines. 
    Our findings highlight several key aspects: 
    Evaluating the generalizability of FNLL algorithms across diverse model architectures is essential; 
    FNLL algorithms need to be robust across different datasets and noise settings; 
    SCE and GCE emerge as the best baselines in balancing accuracy and computational efficiency, recommending their use as primary baselines in FNLL algorithm development; 
    Unlike the case in a centralized setting, symmetric noise can be harder than asymmetric ones under some Non-IID partitions;
    While focusing on denoising proves beneficial at low noise ratios, addressing Non-IID data becomes more critical when dealing with higher noise levels. 
\end{itemize}

\section{Related Works}

\textbf{Noisy Label Learning.} Noisy label learning (NLL) aims to train robust DNNs with noisy labels. Different approaches have been proposed to address this problem in the centralized learning setting, and they can be roughly divided into five categories \cite{9729424}: \begin{enumerate*}[label=\roman*)]
    \item robust architecture \cite{DBLP:journals/tip/YaoWTZSZZ19, DBLP:conf/icml/LeeYLLLS19}; \item robust regularization methods \cite{DBLP:conf/iclr/XiaL00WGC21, DBLP:conf/iclr/ZhangCDL18}, \item robust loss function design \cite{wang2019symmetric, DBLP:conf/icml/MaH00E020}, \item loss adjustment methods \cite{patrini2017making, DBLP:conf/icml/ArazoOAOM19}, and \item sample selection methods \cite{DBLP:conf/iclr/LiSH20, han2018co}. 
\end{enumerate*}
We refer readers to related NLL survey papers for more details (e.g., \cite{9729424}). In \texttt{FedNoisy}, we provide at least one method from each of the last four categories, combining them with FL algorithms as baselines for FNLL research. We plan to support more methods from different categories of NLL in the future.

\textbf{Federated Noisy Label Learning.} Despite diverse research devoted to Non-IID problems and communication efficiencies, most assume that the training datasets only contain clean labels. Some works have been trying to alleviate the label noise in FL recently, which mainly follow three lines: 
\begin{enumerate*}[label=\roman*)]
\item client selection, \item client reweighting, and \item sample selection.
\end{enumerate*}
Client selection methods \cite{DBLP:series/lncs/0013RZJS20, DBLP:journals/corr/abs-2205-10110, DBLP:conf/cvpr/XuCQC22} select clients with high-quality labels based on different evaluation metrics and only allow selected clients to participate FL training rounds. 
Client reweighting methods \cite{DBLP:conf/cvpr/FangY22, DBLP:journals/corr/abs-2001-11359, DBLP:conf/cvpr/XuCQC22} evaluate each client based on the quality of local datasets and assign lower weighting scores to noisy clients during aggregation without abandoning them. 
Sample selection methods evaluate each sample on clients, for example, per-sample loss, and exclude noisy samples for later training  \cite{DBLP:conf/cvpr/XuCQC22, DBLP:conf/icpr/Tuor0KLL20, DBLP:journals/expert/YangPBK22}. 
In addition to the previous mainstream approaches, researchers will reuse the low-quality samples via pseudo-labeling for further performance improvement \cite{DBLP:conf/cvpr/XuCQC22, DBLP:journals/expert/YangPBK22, DBLP:journals/corr/abs-2205-10110}.

\textbf{Federated Learning Benchmark.} Standardized benchmark empowers researchers to follow previous research with less effort and facilitates the validation of research ideas. 
There are several benchmark works in the field of FL, including standardized simulation environments \cite{DBLP:journals/corr/abs-2007-14390, DBLP:journals/corr/abs-2007-13518, JMLR:v24:22-0440}, heterogeneous data distribution \cite{DBLP:conf/icde/LiDCH22, DBLP:journals/corr/abs-1812-01097}, and specific applications \cite{DBLP:conf/nips/TerrailACGHLMMM22, DBLP:conf/nips/SongGT22, DBLP:journals/corr/abs-2104-07145}. 
However, to the best of our knowledge, no existing benchmark is proposed for FL with noisy labels, which leads to the non-standardized simulation pipeline and unfair comparison among related research. 
The proposed \texttt{FedNoisy} bridges the gap between the FL community and existing CNLL research and provides a roadmap for the experiment setting for future FNLL works.

\section{Problem Formulation}
\label{sec:background_problem_formulation}

We begin with formal definitions of federated learning (FL) and noisy label learning (NLL). Then, we formulate the problem of federated noisy label learning (FNLL).

\textbf{Federated Learning} (FL) collaboratively learns a model among multiple participants without directed data sharing \cite{DBLP:conf/aistats/McMahanMRHA17}. 
Here we consider a representation FL setting: $C$-class classification in a horizontal federated learning system with a central server. 
Let $\mathcal{D} = \{ (x^{n}, y^{n}) \}_{n=1}^N$ denote the global training dataset, where $x^n \in \mathcal{X}$ denotes a sample, and $y^n \in \mathcal{Y} = [C]$ is its ground-truth label. 
Suppose there are $K$ clients in the federated system splitting $\mathcal{D}$ without overlap, and for each client $k$ ($k=1,\ldots,K$), there is a local dataset $\mathcal{D}_k = \{ (x_k^n, y_k^n) \}_{n=1}^{N_k}$ such that $N = \sum_{k=1}^K{N_k}$.
% where $N_k = |\mathcal{D}_k|$ is the total sample number of dataset $\mathcal{D}_k$. The total sample number of global dataset $\mathcal{D}$ is $|\mathcal{D}| = N = \sum_{k=1}^K{N_k}$. 
In typical FL problems, the distribution of $\mathcal{D}_k$ varies for different clients and is called Non-IID.

Denote the global model parameter at round $t$ as $w^t$, and the local one of client $k$ as $w^t_k$. 
% For better generalization, we formalize the goal of the task as to find the global model parameter $w$ that minimizes the global objective same as that in \cite{DBLP:conf/icml/00050BS21}:
Formally, FL aims to optimize a global model $w$ by
\begin{equation}
\label{eq:global-obj}
    \min_{w} \mathcal{G}(F_1(w), \ldots, F_K(w)),
\end{equation}
where $F_k(w)$ is the local objective for client $k$, and $\mathcal{G}$ is a function that combines local objectives together. 
$\mathcal{G}(\cdot)$ can be simple weighted average of $\{ F_k(w) \}_{k \in [K]}$ as in FedAvg \cite{DBLP:conf/aistats/McMahanMRHA17}.
% or more sophisticated formulations with extra regularization term and Min-Max optimization \cite{NEURIPS2021_82599a4e, DBLP:journals/corr/abs-2006-11489, DBLP:conf/icml/QuLDLTL22}. 
Local objective $F_k(\cdot)$ can be the empirical risk on the local dataset $\mathcal{D}_k$:
\begin{equation}
    F_k(w) = \mathbb{E}_{(x, y) \sim \mathcal{D}_k}\left[\ell(y, f_k(x; w)) \right], \label{eq:local_train}
\end{equation}
where $\ell$ is a certain loss function, for example, cross-entropy loss, and $f_k(x; w)$ is the local prediction on sample $x$ parameterized by $w$ on client $k$. 
% Extra regularizers can also be plugged into local objective $F_k(\cdot)$ for various learning purposes \cite{DBLP:conf/icml/00050BS21, DBLP:conf/iclr/JeongYYH21}. 
At each communication round $t$, the central server selects a group of clients $\mathcal{S}_t$ and broadcasts global model $w^{t-1}$ to them. 
Then each selected client performs local training in Eq.~(\ref{eq:local_train}) to update $w^t_k$, and sends the local model back to the server. 
The central server then aggregates the local models based on the design of Eq.~(\ref{eq:global-obj}) to obtain the global model $w^t$ and repeats the learning procedure for $T$ rounds.

\textbf{Noisy Label Learning} aims to mitigate the performance drops caused by noised training labels.
% Given a $C$-class dataset with noisy labels as $\tilde{\mathcal{D}} = \{(x^n, \tilde{y}^n)\}^N_{n=1}$, with $\tilde{y}^n \in \mathcal{Y} = [C] = \{1, \ldots, C\}$ denoting annotated label (can be incorrect), and we use $y^n$ to denote corresponding clean label. 
% According to the generation process, the label noise can be roughly divided into two categories.
% \emph{instance-independent} and \emph{instance-dependent}. 
% \textbf{1) Instance-independent noise.} 
The poor label qualities often arise from limited annotation resources where many labels are not fully examined in practice.
Given a well-labeled dataset, the corrupted training dataset is forged by randomly changing true labels following a predefined distribution and then a model is trained on it and evaluated on a clean test set.
The corrupted label $\tilde y$ hinges on the ground-truth label $y$ while independent of its sample $x$. 
% i.e., the probability of the true label $i$ being flipped into a noised label $j$ is $P(\tilde{y} = j \mid y=i, x) = P(\tilde{y}=j \mid y=i)$. 
The noise transition matrix $\mathbf{T} \in [0,1]^{C \times C}$ denotes all possible flipping probability between classes, that is, $\mathbf{T}_{ij} = P(\tilde{y}=j \mid y=i)$. 
% The most typical types of this category
Given a constant noise ratio $\varepsilon \in [0,1]$, representative instantiations of such noise could \emph{symmetric noise} or \emph{asymmetric one}. 
(1) In the symmetric setting, clean labels are uniformly corrupted into other labels with equal probability, i.e., $\forall_{i \neq j} \mathbf{T}_{ij} = \frac{\varepsilon}{C-1}$ and $\forall_{i=j} \mathbf{T}_{ij}=1-\varepsilon$. 
(2) In the asymmetric setting, the label will be only flipped to a particular label, i.e., $\forall_{i=j} \mathbf{T}_{ij} = 1 - \varepsilon$, $\exists_{c \neq i}\mathbf{T}_{ic} = \varepsilon$, and $\forall_{j\neq i, j\neq c} \mathbf{T}_{ij} = 0$. Such asymmetric noise is also called pairwise flipping. 
% \textbf{2) Instance-dependent noise.}
In some literature~\cite{DBLP:conf/cvpr/XiaoXYHW15}, the corruption probability can be formulated as $P(\tilde{y}=j \mid y=i, x)$, namely instance-dependent noise. 
% In our benchmark, FL label corruption processes include symmetric noise and asymmetric noise. 
In this paper, we only consider symmetric noise, asymmetric noise following common practice, and also include dataset with real-world noise.

\textbf{Federated Noisy Label Learning.} 
% \comm{I don't understand: In this setting, we assume clients can only collect datasets with noisy labels in a federated system.} 
Distributed clients in a federated system have been granted the freedom to choose training data and keep the data confidential to the server. 
As such, the label quality cannot be sanitized by the trustworthy server using typical noise label treatments.
To simulate the scenario, we let $\tilde{\mathcal{D}} = \{ (x^n, \tilde{y}^n) \}_{n=1}^N$ be the noisy global dataset in the federation. And each client holds a corrupted local dataset $\tilde{\mathcal{D}}_k = \{(x^n_k, \tilde{y}^n_k)\}_{n=1}^{N_k}$  instead of $\mathcal{D}_k$. Then the local objective $F_k(\cdot)$ becomes:
\begin{equation}
    F_k(w) = \mathbb{E}_{(x, \tilde{y}) \sim \tilde{\mathcal{D}}_k}\left[\ell(\tilde{y}, f_k(x; w)) \right],
\end{equation}
which is the empirical risk on corrupted local dataset $\tilde{\mathcal{D}}_k$. In this study, we use vanilla FedAvg for server aggregation and our benchmark can be easily extended to other aggregation approaches.

\section{Benchmark Design}

In this section, we introduce federated noisy-label settings and code resources provided by \texttt{FedNoisy}.
First, we provide fine-grained and flexible federated noisy-label configurations, with which users can evaluate their method in multiple settings.
We provide datasets with well-designed codes that can be used for simulating the federated heterogeneity.
In our benchmark, 8 baseline methods have been implemented for method comparisons.

% We first introduce the overall federated label noise schemes, then the data partition schemes used in federated noise schemes.

\subsection{Label-Noise Simulations for Heterogeneous Clients}

In FNLL, Non-IID and noisy labels are concurrent and intervened problems when training a robust global model. 
Non-IID means that the distributions of samples are different among clients. 
Such distribution shift can also happen to noisy labels in FL, i.e., the distribution of noisy labels on each client can also differ, which we refer to as \emph{heterogeneous label noise}.
Motivated by this, we propose three different federated noise schemes to represent different generation processes of label noise in FL: globalized noise scene for homogeneous label noise, localized noise scene for heterogeneous label noise, and real-world noise scene for real-world setting. 

\textbf{Globalized noise scene.}
The globalized noise scene echoes the generation process of noisy labels where the whole federated system follows a specific unified distribution. 
In other words, the noisy labels on clients are independent and identically distributed. And globalized noise scene can also be called \emph{homogeneous label noise}.

\begin{figure*}
	\centering
        \includegraphics[width=0.6\textwidth]{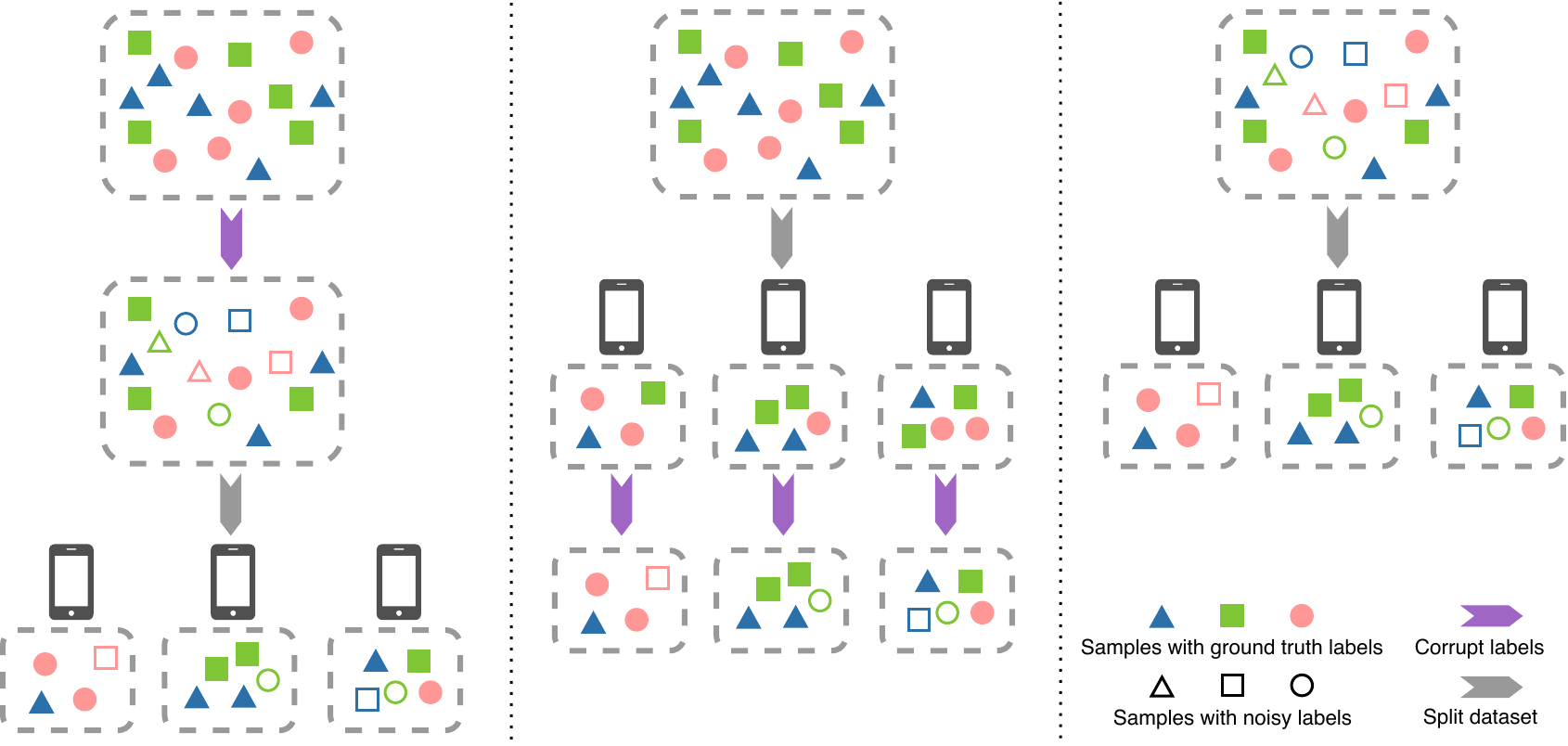}
	\caption{Federated noise scenes provided in \texttt{FedNoisy}. Left: \emph{globalized noise}; middle: \emph{localized noise}; right: \emph{real-word noise}. Globalized noise follows a constant noise ratio $\varepsilon_{global}$ in label corruption, while localized noise draws localized noise ratio $\varepsilon_{k}$ from a uniform distribution on each client $k$.}
	\label{fig:fednoisy-noise-scenes}
\end{figure*}

The global dataset in this scene follows noise distributions with a fixed global noise ratio $\varepsilon_{global}$. 
To be more concrete, a global noise transformation matrix $\mathbf{T}_{global} \in [0,1]^{C \times C}$ will be applied on the clean global dataset $\mathcal{D}$ to flip labels based on global noise ratio $\varepsilon_{global}$ for symmetric/asymmetric noise. 
Note $C$ is the total class number in the FL. 
Then corrupted global dataset $\tilde{\mathcal{D}}$ will be split into $K$ local noisy datasets $\{ \tilde{\mathcal{D}}_1, \ldots, \tilde{\mathcal{D}}_K \}$ in IID/non-IID fashion. The left one in Figure~\ref{fig:fednoisy-noise-scenes} shows the generation process of the globalized noise scene.

\textbf{Localized noise scene.}
In real-world federated systems, label noise may be attributed to clients' local data collection process, which means each client has an independent label corruption process with local label noise distribution. 
For example, hospitals with limited medical resources are more likely to have higher misdiagnosis rates~\cite{zhang2020intelligent}. 
To simulate this, we propose \emph{localized noise scene}, in which different clients have different local noise ratios of noisy label distribution. And it can also be called \emph{heterogeneous label noise}.

In localized noise scene, a clean global dataset $\mathcal{D}$ is first split into $K$ local clean datasets $\{\mathcal{D}_1, \ldots, \mathcal{D}_K \}$, following IID/non-IID fashion. 
For each client $k$, the local noise distribution is synthesized using a local noise transformation matrix $\mathbf{T}_k \in [0,1]^{C_k \times C_k}$ generated from local noise ratio $\varepsilon_k$, where $\varepsilon_k$ is drawn from a uniform distribution $\mathcal{U}(\varepsilon_{min}, \varepsilon_{max})$ with $\varepsilon_{min}, \varepsilon_{max} \in [0,1]$, and $C_k$ is the total number of classes on clean local dataset $\mathcal{D}_k$. 
We flip sample labels of the local clean dataset $\mathcal{D}_k$ using $\mathbf{T}_k$ for symmetric/asymmetric noise and obtain noisy local dataset $\tilde{\mathcal{D}}_k$ for each client $k$. 
Also, the label noising process on each client only allows label flipping among classes in the current local client. 
To be more concrete, given an FL system with sample labels of $\{1, 2, 3, 4, 5\}$ in the global dataset, if a particular client only holds a local dataset with class label $\{1, 3, 5\}$, then there are no chances for the existence of noisy samples with class label $\{2,4\}$ on this client after label corruption. 
This constraint of the localized scene closely resembles many real-world cases: a dentist may misdiagnose different dental diseases while unlikely to diagnose lung cancer. 
The middle part of Figure~\ref{fig:fednoisy-noise-scenes} shows the generation process of the localized noise scene.

\textbf{Real-world noise scene.} 
Unlike globalized and localized noise scenes that only contain synthetic label noise, real-world noise scene evaluates against datasets inherently with mislabeled samples. 
Here we use two popular real-world noisy datasets in CNLL studies \cite{9729424}, that is, \emph{Clothing1M}~\cite{DBLP:conf/cvpr/XiaoXYHW15} and \emph{WebVision}~\cite{DBLP:journals/corr/abs-1708-02862}~\footnote{We use the first 50 classes of the Google subset from WebVision 1.0 resized version as noisy global dataset $\tilde{\mathcal{D}}$ in our benchmark.}, as summarized in Table~\ref{tab:noisy-dataset}. 
% \textbf{WebVision} \cite{DBLP:journals/corr/abs-1708-02862} is a real-world image dataset crawled from Flickr and Google. The labels are enquiring text generated from $1000$ concepts in ImageNet ILSVRC12 \cite{DBLP:conf/nips/KrizhevskySH12}, rather than human-annotated ones. Following the setting in previous centralized noisy label learning researches \cite{DBLP:conf/icml/JiangZLLF18, DBLP:conf/icml/ChenLCZ19, DBLP:conf/iclr/LiSH20}, we use the first 50 classes of the Google subset from WebVision 1.0 resized version~\footnote{\url{https://data.vision.ee.ethz.ch/cvl/webvision/download.html}} as noisy global dataset $\tilde{\mathcal{D}}$ in our benchmark.
In the real-world noise scene, we consider directly partitioning the noisy global dataset $\tilde{\mathcal{D}}$ into $\{\tilde{\mathcal{D}}_1, \ldots, \tilde{\mathcal{D}}_K\}$ via IID/non-IID fashion to create an FL setting with real-world label noise. The right part of Figure~\ref{fig:fednoisy-noise-scenes} shows the generation process of the real-world noise scene.

% \begin{figure}[h!]
% 	\centering
%         \includegraphics[height=4cm]{imgs/real-world-noise.pdf}
% 	\caption{Real-world noise scene}
% 	\label{fig:real-noise}
% \end{figure}

\subsection{Datasets and Data Heterogeneity Simulation}
\label{sec:datasets-data-heterogeneity}

As summarized in Table~\ref{tab:noisy-dataset}, \texttt{FedNoisy} serves 5 datasets for variant tasks.
MNIST and SVHN are digit recognition tasks where digits 1 to 10 are included.
CIFAR-10 and CIFAR-100 are popular object recognition datasets with 10 and 100 classes, respectively.
These four datasets are frequently adopted in both FL and CNLL areas. 
We apply globalized and localized noise scenes to them and evaluate them on the clean test set.
Clothing1M is a large-scale clothing image dataset of 14 categories from shopping websites, containing natural label noise at a ratio of 39.46\%. It is a popular dataset widely used in CNLL, and we use it for the real-world noise scene in \texttt{FedNoisy}. 
Following DivideMix \cite{DBLP:conf/iclr/LiSH20}, we randomly sample a class-balanced subset containing 64k images from the original noisy training set as the noisy global dataset $\tilde{\mathcal{D}}$ for training, and report the test performance on 10k clean test set.

\begin{table*}
\centering
\caption{Summary of datasets provided in \texttt{FedNoisy}}
\label{tab:noisy-dataset}
\scalebox{0.78}{
\begin{tabular}{clcccccl} 
\toprule
\multicolumn{1}{c}{\textbf{Noise Scene}}                & \textbf{Dataset} & \textbf{\#Train} & \textbf{\#Validation} & \textbf{\#Test} & \textbf{\#Class} & \textbf{ImageSize}      & \textbf{Noise Ratio (\%)}  \\ 
\midrule
\multirow{4}{*}{\textit{Globalized/Localized}}          & MNIST \cite{lecun1998mnist}            & 60K                 & N/A                   & 10K                & 10                 & $28 \times 28$          & N/A                        \\
                                                         & SVHN \cite{netzer2011reading}             & 73K                 & N/A                   & 26K                & 10                 & $32 \times 32 \times 3$ & N/A                        \\
                                                         & CIFAR-10 \cite{krizhevsky2009learning}         & 50K                 & N/A                   & 10K                & 10                 & $32 \times 32 \times 3$ & N/A                        \\
                                                         & CIFAR-100 \cite{krizhevsky2009learning}        & 50K                 & N/A                   & 10K                & 100                & $32 \times 32 \times 3$ & N/A                        \\
\midrule
\multicolumn{1}{c}{\multirow{2}{*}{\textit{Real-world}}} & Clothing1M \cite{DBLP:conf/cvpr/XiaoXYHW15}       & 1.0M                  & 14K                   & 10K                & 14                 & $224 \times 224 \times 3$        & $\approx 39.46$            \\
\multicolumn{1}{c}{}                                     & WebVision \cite{DBLP:journals/corr/abs-1708-02862}        & 2.4M                & 50K                   & 50K                & 1000               & $256 \times 256 \times 3$        & $\approx 20.00$            \\
\bottomrule
\end{tabular}}
\end{table*}

% We also follow previous arts to simulate the distribution heterogeneity among clients.
Found by massive existing FL research \cite{DBLP:conf/aistats/McMahanMRHA17, DBLP:journals/ftml/KairouzMABBBBCC21, DBLP:conf/icde/LiDCH22}, data heterogeneity is a common yet especially challenging scenario in practice where each client has non-identical-and-independent (non-IID) data distributions compared to others.
Given a data sample $(x, y)$, non-IID indicates the joint probability distribution $\mathcal{P}(x, y)$ is different for different client $i$ and $j$ ($i, j \in [K]$, $i \neq j$), that is, $\mathcal{P}_i(x,y) \neq \mathcal{P}_j(x,y)$. 
In \texttt{FedNoisy}, we enclose three representative non-IID schemes together with the standard IID setting, as follows. 
%By Bayes' theorem, $\mathcal{P}(x, y)$ can be rewritten as $\mathcal{P}(y|x) \mathcal{P}(x)$ and $\mathcal{P}(x|y) \mathcal{P}(y)$. 
\emph{1)~Quantity skew.} Different clients hold different amounts of data, controlled by Dirichlet distribution with parameter $\alpha$, denoted as \texttt{noniid-quantity=$\alpha$}. 
\emph{2)~Dirichlet label skew.} Different clients hold local datasets with different proportions of different classes, controlled by Dirichlet distribution with parameter $\alpha$. We denote this as \texttt{noniid-labeldir=$\alpha$}. 
\emph{3)~Label-quantity skew.} In this Non-IID case, the class heterogeneity is simulated by assigning samples of a specified number of classes to each client. We use \texttt{noniid-\#label=c} to denote this partition given class number \texttt{c}. Detailed descriptions of the data partition are presented in Appendix~\ref{append:data-partition}.

\subsection{Implemented methods}

\begin{table}[h]
\centering
\caption{Summary of baselines in \texttt{FedNoisy}}
\label{tab:baseline}
\scalebox{0.8}{
\begin{tabular}{ccc} 
\toprule
\multicolumn{2}{c}{\textbf{Noisy Label Algorithm}}                                                    \\ 
% \cmidrule(lr){2-2}\cmidrule(lr){3-3}
\textit{Category}                     & \textit{Method}                                               \\ 
\midrule
Robust regularization                 & Mixup~\cite{DBLP:conf/iclr/ZhangCDL18}       \\ 
\midrule
\multirow{3}{*}{Robust loss function} & SCE~\cite{wang2019symmetric}                 \\
 & GCE \cite{zhang2018generalized}              \\
                   & MAE~\cite{DBLP:conf/aaai/GhoshKS17}          \\ 
\midrule
\multirow{3}{*}{Loss adjustment}      & M-DYR-H~\cite{DBLP:conf/icml/ArazoOAOM19}    \\
                                   & M-DYR-S~\cite{DBLP:conf/icml/ArazoOAOM19}    \\
                                    & DM-DYR-SH~\cite{DBLP:conf/icml/ArazoOAOM19}  \\ 
\midrule
Sample selection                      & Co-teaching~\cite{han2018co}                 \\
\midrule
\multirow{1}{*}{FNLL} & FedNoRo~\cite{wu2023fednoro}                 \\
 % & GCE \cite{zhang2018generalized}              \\
 %                   & MAE \cite{DBLP:conf/aaai/GhoshKS17}          \\ 
\bottomrule
\end{tabular}}
\end{table}

In \texttt{FedNoisy}, we integrate representative NLL algorithms in the local client training, which can work seamlessly with most federated learning algorithms.
The list of algorithms is summarized in \ref{tab:baseline}.
Previous NLL methods can be roughly grouped into four categories: \begin{enumerate*}[label=\roman*)]
    \item robust regularization methods, \item robust loss function design, \item loss adjustment methods, and \item sample selection methods. 
\end{enumerate*} 
Though many algorithms have been developed under these principles, they are not extensively used as baselines in existing FNLL research.
The main reason could be a lack of a well-developed codebase, leading to formidable burdens including them for comparisons.
% Thus, researchers may overlook the value of such previous arts in FNLL.
In our benchmark, we tackle the critical challenge by providing a package of ready-to-go codes for these algorithms.
% most recent FNLL works naively combine only state-of-the-art NLL methods with FL algorithms as baselines in experiments.
Importantly, we implement these algorithms mainly on the client side with the least modifications on the global communication.
The decoupled code structure makes these algorithms easy to be extended with more advanced federated aggregations.

% Such prosperous directions facilitate NLL research. 
% However, most recent FNLL works combine only SOTA CNLL methods with FL algorithms as baselines in experiments, most of which only belong to the sample selection method category. 
% These limited baselines hinder the development of FNLL research. 
% We select at least one method from each of last four method categories, aiming to cover a broad scope of NLL research, and combine them with FedAvg by deploying NLL methods into the process of client local training. 
% The baseline algorithms in \texttt{FedNoisy} are listed in Table~\ref{tab:baseline}.

\section{Experiments}

To demonstrate the utility of FNLL data settings proposed by \texttt{FedNoisy}, we conduct extensive experiments and present the main results for characteristics of these settings under different datasets.  

\subsection{Benchmark performance}
\label{sec:bench-performance}

This section presents the performance for 20 benchmark settings in \texttt{FedNoisy}. 
We compare the test accuracy of 20 benchmark settings in \texttt{FedNoisy} as an overview of the challenges in FNLL.
In the experiment, we select some default hyper-parameters: $\varepsilon_{global}=0.4$ for globalized noise.
For localized noise, we use $\varepsilon_k \sim \mathcal{U}(0.3, 0.5)$  (that is, $\varepsilon_{local}=0.4$, $\sigma=0.1$).
The parameters of localized noise yield a fair comparison between localized and globalized settings with similar overall noisy samples, which is shown in Appendix~\ref{append:fair-comparison-local-global}). 
For data partition, we choose \texttt{iid}, \texttt{noniid-\#label=3}, \texttt{noniid-labeldir=0.1} and \texttt{noniid-quantity=0.1} as the default settings. 
All analyses are based on the results of CIFAR-10 on FedAvg of 10 clients using VGG-16 \cite{DBLP:journals/corr/SimonyanZ14a} with 500 communication rounds and 5 local epochs. 
We report the averaged test accuracy over the last 10 rounds, and each experiment is repeated 3 times with different random seeds. 
More details on experiments can be found in Appendix~\ref{append:experiment-detail}. 
Table~\ref{tab:cifar10-10-clients-fedavg} presents the complete results for 20 default settings. 
Results for the same experiment setting for more datasets can be found in Appendix~\ref{append:other-dataset-results}. 

\begin{table*}
\centering
\caption{Test accuracy of CIAFR-10 on 10 clients using FedAvg with VGG16. Significant gaps exist between different noise schemes, especially under heterogeneous data partitions.}
\label{tab:cifar10-10-clients-fedavg}
\scalebox{0.78}{
\begin{tabular}{ccccccc} 
\toprule
\multirow{2}{*}{}           & \multirow{2}{*}{\textbf{Noise mode}} & \multirow{2}{*}{\textbf{Noise ratio}}    & \multicolumn{4}{c}{\textbf{Data partition}}                                    \\ 
\cmidrule(lr){4-4}\cmidrule(lr){5-5}\cmidrule(lr){6-6}\cmidrule(lr){7-7}
                            &                                      &                                          & \texttt{iid}               & \texttt{noniid-\#label}    & \texttt{noniid-labeldir}   & \texttt{noniid-quantity}    \\ 
\midrule
                            & Clean                                & $0.0$                                    &  $90.83 \pm 0.17$ & $73.09 \pm 4.33$ & $78.74 \pm 3.56$ & $91.26 \pm 0.41$  \\ 
\midrule
\multirow{2}{*}{Globalized} & Sym.                                 & \multirow{2}{*}{$0.4$}                   &  $65.08 \pm 0.38$ & $33.04 \pm 1.87$ & $30.43 \pm 6.30$ & $64.28 \pm 3.07$  \\ 
\cmidrule(lr){2-2}
                            & Asym.                                &                                          &  $57.85 \pm 0.59$ & $44.06 \pm 3.65$ & $42.02 \pm 5.51$ & $57.13 \pm 0.79$ \\ 
\midrule
\multirow{2}{*}{Localized}  & Sym.                                 & \multirow{2}{*}{$\mathcal{U}(0.3, 0.5)$} &  $65.51 \pm 0.80$ & $17.25 \pm 1.83$ & $26.09 \pm 8.68$ & $62.89 \pm 4.76$ \\ 
\cmidrule(lr){2-2}
                            & Asym.                                &                                          &  $58.42 \pm 1.50$ & $26.51 \pm 2.90$ & $39.5 \pm 6.67$  & $54.84 \pm 3.07$ \\
\bottomrule
\end{tabular}}
\end{table*}

\paragraph{Basic observations.} From the results in Table~\ref{tab:cifar10-10-clients-fedavg}, we have the following basic observations:
\emph{1) Class imbalance should be considered together with label noise.} In both the clean and the globalized noise settings, the performance difference between \texttt{iid} and \texttt{noniid-quantity} is limited. 
In contrast, the difference between \texttt{iid} and \texttt{noniid-labeldir}/\texttt{noniid-\#label} is significant under both globalized and localized noise, which indicates that label distribution skew has a large effect on performance degradation in FNLL. 
We suggest that researchers should consider class imbalance and label noise together in FNLL from this. 
\emph{2) Local noise patterns are a new challenge.} 
By comparing performances between globalized and localized noise, we found localized noise is harder than globalized noise in all Non-IID partitions, while not in the IID partition. 
This indicates that Non-IID (quantity imbalance and class imbalance) can exaggerate the effect of heterogeneous noise distributions in FL. 
And knowledge of different noise patterns on different clients cannot be aggregated by simple weighted averaging for a global shared model. 
This encourages researchers in FNLL to consider local adaption for noise heterogeneity as well as more sophisticated designs for global aggregation.

\subsection{Baseline Performance}

\begin{figure*}
    \centering
    \includegraphics[width=0.7\textwidth]{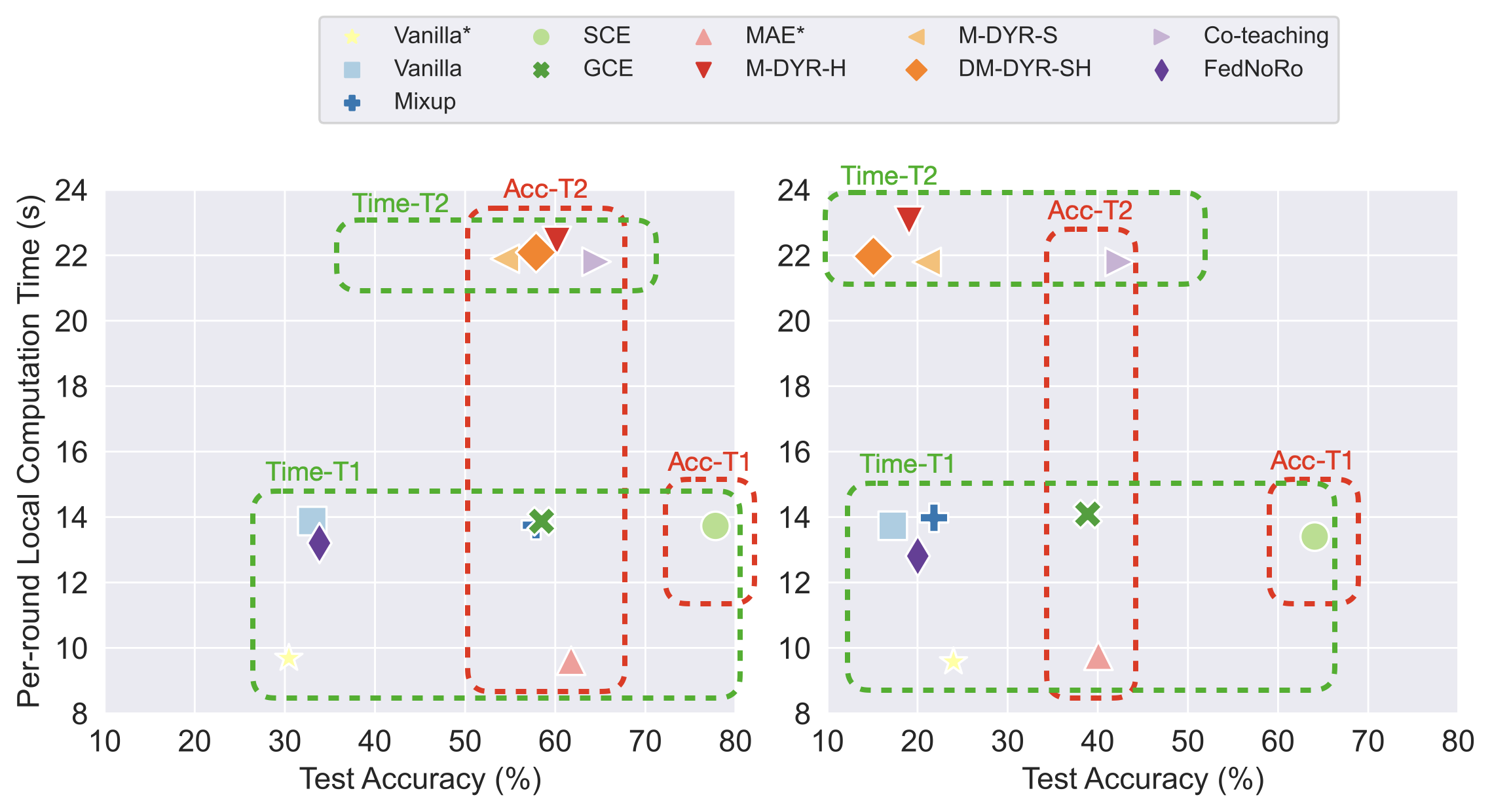}
    \caption{The trade-off between accuracy and per-round local computation time on CIFAR-10 on 10 clients under \texttt{noniid-\#label=3} partition with symmetric noise. From left to right: globalized noise, localized noise. Baselines with \texttt{*} are using 8-layer CNN following \cite{DBLP:conf/icml/MaH00E020}, otherwise with default VGG16. 
    % Red and green dash boxes are used to categorize the baselines into several tiers for both test accuracy and per-round time, respectively. \textit{T1} denotes tier 1, and \textit{T2} denotes tier 2.
    }
    \label{fig:baseline-results-cifar10}
\end{figure*}

\begin{figure}[!ht]
    \centering
    \includegraphics[width=0.35\textwidth]{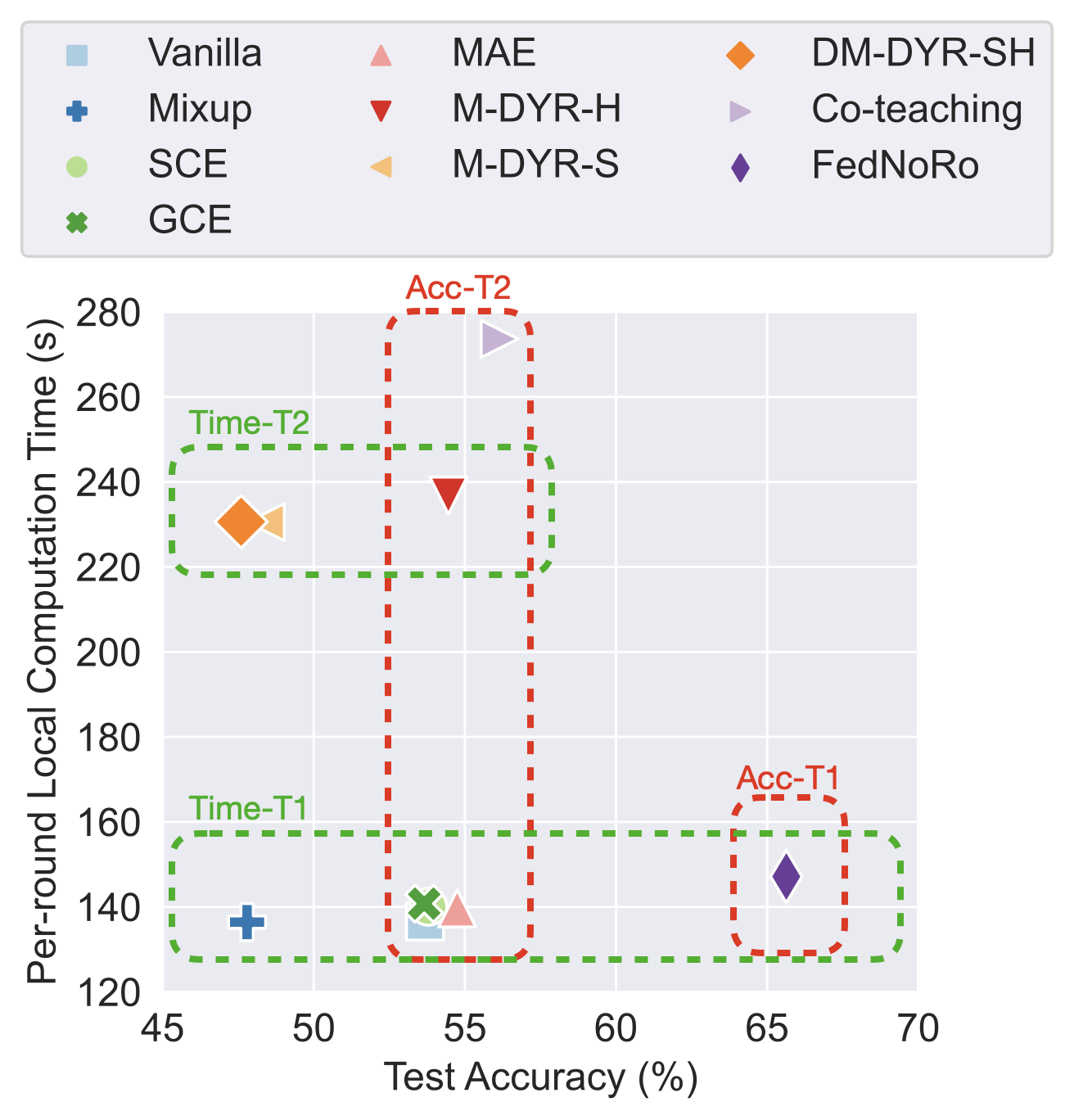}
    \caption{The trade-off between accuracy and per-round local computation time on Clothing1M on 10 clients under \texttt{noniid-\#label=5} partition with real noise. Red and green dash boxes are used to categorize the baselines into several tiers for both test accuracy and per-round time, respectively. \textit{T1} denotes tier 1, and \textit{T2} denotes tier 2.}
    \label{fig:baseline-results-clothing1m}
\end{figure}

We evaluate all baselines using MNIST, CIFAR-10, and Clothing1M under \texttt{noniid-\#label} setting. 
We aim to explore the trade-off between local computation cost and model performance.
The scatter visualizations for CIFAR-10 and Clothing1M are shown in~\autoref{fig:baseline-results-cifar10} and \autoref{fig:baseline-results-clothing1m}, and the result for MNIST is shown in~\autoref{fig:baseline-results-mnist} of Appendix G for space limit. 
We categorize the baselines based on their ranking into tier 1 (\textit{T1}), tier 2 (\textit{T2}), and others for accuracy and computation time, respectively. We use \textcolor{red}{red dashed} boxes for the accuracy tiers and \textcolor{dashboxgreen}{green dashed} boxes for time.

\paragraph{Generalizability over different model architectures.} 
For CIFAR-10, we find that the MAE yields reasonable performance with an 8-layer CNN, but fails to converge on VGG16. 
This inconsistency underscores the need for researchers to consider the generalizability of the FNLL across various model architectures.

\paragraph{Baseline robustness across datasets and noise settings.} By comparing \autoref{fig:baseline-results-mnist}, \autoref{fig:baseline-results-cifar10}, and \autoref{fig:baseline-results-clothing1m}, we find that the performance rankings of baselines exhibit variability across different datasets as well as noise settings, indicating a lack of robustness. 
For example, M-DYR-H is in accuracy \textit{T2} for globalized noise, while only achieving only comparable with naive baseline under localized noise on CIFAR-10, as shown in~\autoref{fig:baseline-results-cifar10}. FedNoRo is in accuracy \textit{T1} on Clothing1M as shown in~\autoref{fig:baseline-results-clothing1m}, while is even worse than the vanilla baseline for both CIFAR-10 and MNIST. This indicates that researchers should pay more attention to the algorithms' robustness over datasets and noise settings when working on FNLL.

\paragraph{Accuracy and computation trade-off.} Based on results in \autoref{fig:baseline-results-mnist}, \autoref{fig:baseline-results-cifar10}, and \autoref{fig:baseline-results-clothing1m}, we assign points to baselines based on their accuracy/computation tier: 1 point for tier 1 (\textit{T1}) reflecting either accuracy or local computation time, 2 points for tier 2 (\textit{T2}), and 4 points for the remaining.
We aggregate these points across all datasets to derive a final score for each baseline, assessing the trade-off between accuracy and computational efficiency.
A lower score indicates superior overall accuracy or lower overall computational costs.
As shown in~\autoref{tab:acc-comp-summary}, SCE and GCE achieve the best trade-off between accuracy and computation cost~\footnote{We exclude MAE as the best baseline due to its limited generalizability across different model architectures.}.
Consequently, we recommend that researchers consider both accuracy and computational cost, and adopt SCE and GCE as strong baselines for both criteria when developing FNLL algorithms.

\begin{table}[h]
    \centering
    \begin{tabular}{lccc}
\toprule
\textbf{Algorithm} & \textbf{Accuracy}$\downarrow$ & \textbf{Computation}$\downarrow$ \\
\hline
Vanilla & 18 & 5 \\
Mixup & 12 & 5 \\
\cellcolor{LightCyan} SCE & \cellcolor{LightCyan} 7 & \cellcolor{LightCyan} 5 \\
\cellcolor{LightCyan} GCE & \cellcolor{LightCyan} 9 & \cellcolor{LightCyan} 5 \\
MAE$^*$ & 8 & 5 \\
M-DYR-H & 10 & 14 \\
M-DYR-S & 12 & 14 \\
M-DYR-SH & 12 & 14 \\
Co-teaching & 8 & 12 \\
FedNoRo & 17 & 5 \\
\bottomrule
\end{tabular}
    \caption{Accuracy and time trade-off scores for different baselines over datasets MNIST, CIFAR-10, and Clothing1M. The blue background indicates the baselines with the best trade-off between accuracy and computation cost. $^*$ indicates the baseline has trouble with convergence when using different model architectures.} 
    \label{tab:acc-comp-summary}
\end{table}

% \begin{figure*}
%     \centering
%     \includegraphics[width=0.7\textwidth]{imgs/mnist-10-clients-noniid_label3-comp-cost-vs-acc-global-local.pdf}
%     \caption{The trade-off between accuracy and per-round local computation time on MNIST on 10 clients under \texttt{noniid-\#label=3} partition with symmetric noise. From left to right: globalized noise, localized noise.}
%     \label{fig:baseline-results-mnist}
% \end{figure*}

\subsection{Fine-grained Ablation Studies of FNLL}

\begin{figure*}
    \centering
    \includegraphics[width=0.8\textwidth]{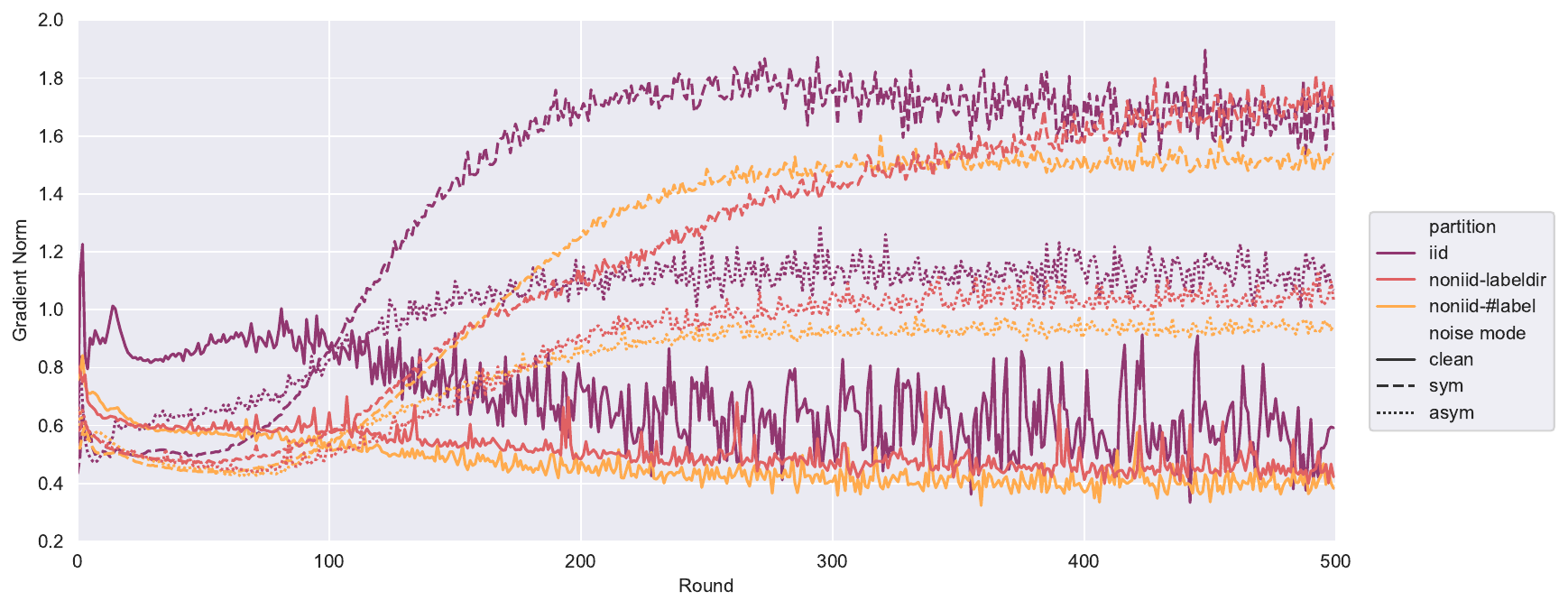}
    \caption{Gradient norm of global model for clean and globalized noise $\varepsilon_{global}=0.4$ settings.}
    \label{fig:grad-norm-global}
\end{figure*}

In the later experiment, to investigate the characteristics of FNLL settings, we perform FedAvg on \texttt{iid}, \texttt{noniid-labeldir},  \texttt{noniid-} \texttt{\#label} partition under different noise settings with different noise ratios for a systematic evaluation. 
We aim to provide insights into the following questions involved in FNLL: 
\emph{1) Is asymmetric noise still harder than symmetric ones in FL as in CNLL?} 
\emph{2) How does label noise affect the model performance with Non-IID? Does the effect enlarge as the noise ratio increases?} 
\emph{3) Does model performance react similarly to label noise under different partitions?} 
\emph{4) How does gradient norm act upon label noise in FL?}

\begin{figure}[!ht]
    \centering
        \includegraphics[width=1\columnwidth]{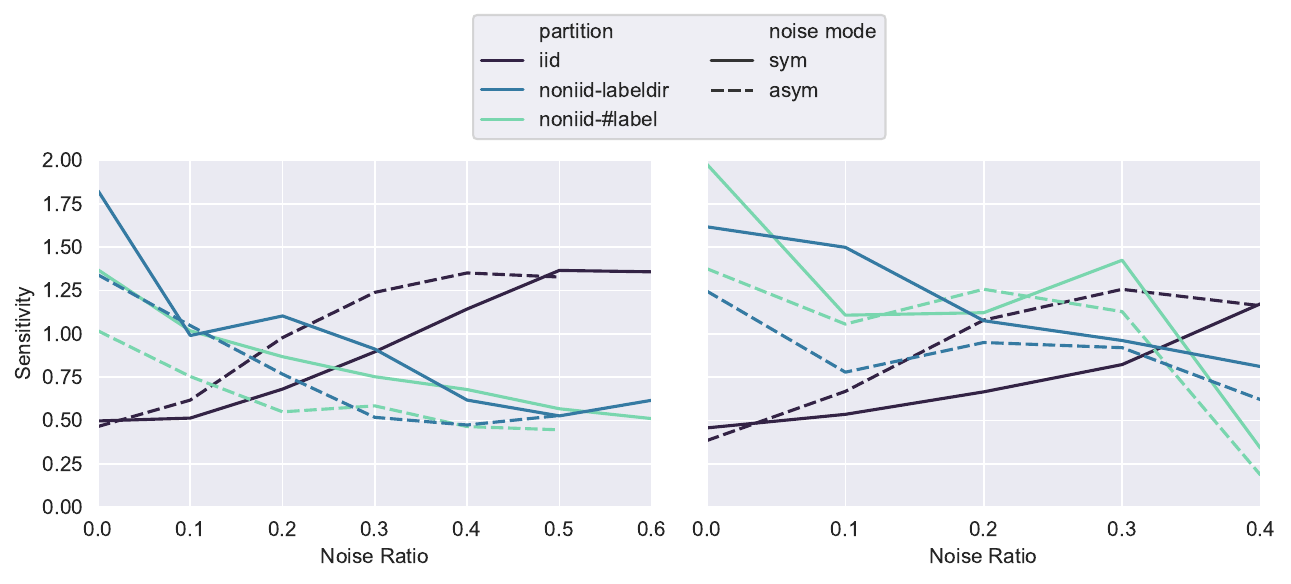}
	\caption{Sensitivity on different noise ratios on CIFAR-10 with 10 clients. The $x$-axis noise ratio is $\varepsilon_{global}$ for globalized noise, and $\varepsilon_{local}$ for localized noise $\mathcal{U}(\varepsilon_{local}-0.1, \varepsilon_{local}+0.1)$. From left to right: globalized noise, localized noise.}
	\label{fig:diff-noise-ratio}
\end{figure}

\sloppy For globalized noise, we choose the global noise ratio $\varepsilon_{global}$ from $\{0.1, 0.2, 0.3, 0.4, 0.5, 0.6, 0.7\}$ for symmetric noise, while $\{0.1, 0.2, 0.3, 0.4, 0.5, 0.6\}$ for asymmetric noise, while keeping other learning setting same as default settings. 
For localized noise, choose $\varepsilon_{local}$ from $\{0.1, 0.2, 0.3, 0.4, 0.5\}$ and $\sigma=0.1$ for both symmetric and asymmetric noise. 
And to make sure localized settings with large noise ratios converge, we train FedAvg for 700 rounds while keeping other learning settings default. 
Figure~\ref{fig:diff-noise-ratio} shows the curves of test accuracy versus noise ratio for both globalized and localized noise settings. 
Complete results are in Appendix \ref{append:cifar10-diff-noise}.

% \begin{figure}[h!]
% 	\centering
%         \includegraphics[width=0.8\textwidth]{imgs/cifar10-10clients-diff-noise-ratio.pdf}
% 	\caption{Accuracy for different noise ratios on CIFAR-10 with 10 clients. The $x$-axis noise ratio is $\varepsilon_{global}$ for globalized noise, and $\varepsilon_{local}$ for localized noise $\mathcal{U}(\varepsilon_{local}-0.1, \varepsilon_{local}+0.1)$. From left to right: globalized noise, localized noise. }
% 	\label{fig:diff-noise-ratio}
% \end{figure}

% \begin{figure}[!ht]
%     \begin{minipage}{0.49\linewidth}
%     \includegraphics[width=\textwidth]{imgs/cifar10-10clients-diff-noise-ratio.pdf}
%     \caption{Accuracy for different noise ratios on CIFAR-10 with 10 clients. The $x$-axis noise ratio is $\varepsilon_{global}$ for globalized noise, and $\varepsilon_{local}$ for localized noise $\mathcal{U}(\varepsilon_{local}-0.1, \varepsilon_{local}+0.1)$. From left to right: globalized noise, localized noise.}
%     \label{fig:diff-noise-ratio}
%     \end{minipage}
%     \hfill
%     \begin{minipage}{0.49\linewidth}
%       \includegraphics[width=\textwidth]{imgs/cifar10-10clients-global-local-sensitive-noise-ratio.pdf}
%     \caption{Sensitivity on different noise ratios on CIFAR-10 with 10 clients. The $x$-axis noise ratio is $\varepsilon_{global}$ for globalized noise, and $\varepsilon_{local}$ for localized noise $\mathcal{U}(\varepsilon_{local}-0.1, \varepsilon_{local}+0.1)$. From left to right: globalized noise, localized noise.}
%     \label{fig:sensitive-noise}
%     \end{minipage}
%   \end{figure}

\begin{figure}[!ht]
    \centering
        \includegraphics[width=1\columnwidth]{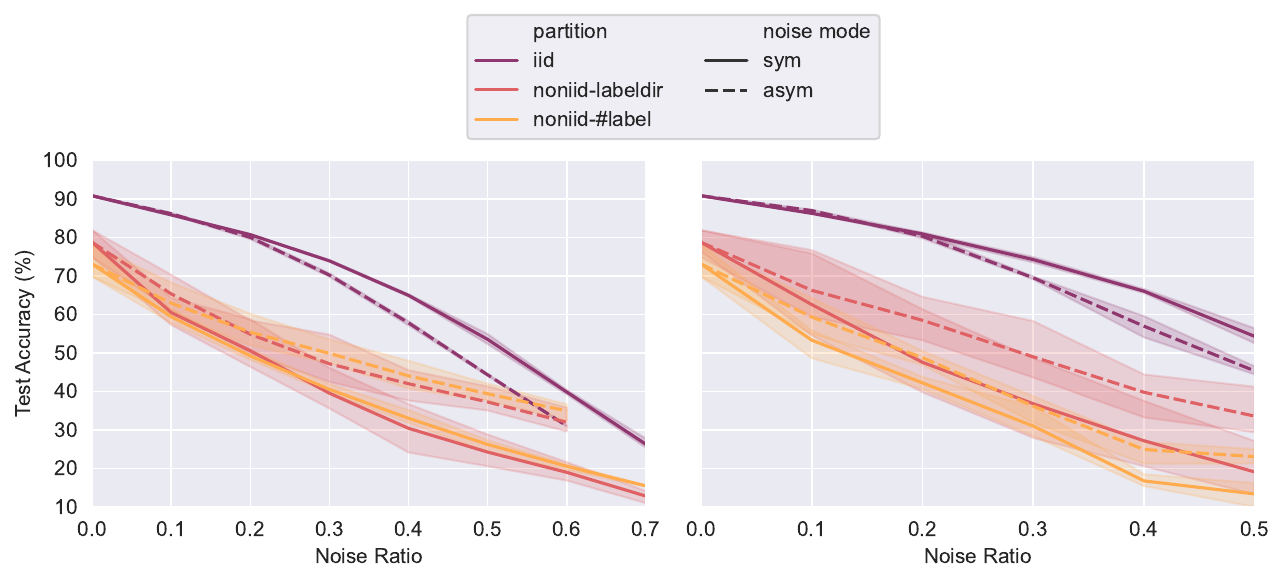}
	\caption{Accuracy for different noise ratios on CIFAR-10 with 10 clients. The $x$-axis noise ratio is $\varepsilon_{global}$ for globalized noise, and $\varepsilon_{local}$ for localized noise $\mathcal{U}(\varepsilon_{local}-0.1, \varepsilon_{local}+0.1)$. From left to right: globalized noise, localized noise.}
	\label{fig:sensitive-noise}
\end{figure}

\textbf{1) Symmetric noise is harder than asymmetric with Non-IID.} 
In the centralized setting, it has been found that asymmetric label noise is harder than symmetric noise, especially when the noise ratio is larger than $0.5$, considering each class has more correct samples to learn from under symmetric noise \cite{han2018co, DBLP:conf/iclr/LiSH20}. 
However, in FNLL, the conclusion depends on the partition type.
As shown in Figure~\ref{fig:diff-noise-ratio}, in both globalized and localized settings, performance with symmetric noise tends to become better than that with asymmetric noise when the noise ratio is larger than $0.2$ under IID partition (class-balanced on each client in this experiment), and the performance gap increases as noise ratio grows. 
While in Non-IID which class-imbalance exists (i.e., \texttt{noniid-labeldir} and \texttt{noniid-\#label}), symmetric noise is more difficult than asymmetric noise even with a small noise ratio. 
And this performance gap also becomes larger as the noise ratio grows. 
This intriguing phenomenon might be attributed to the class imbalance problem in these label distribution skew scenes, which encourages researchers to alleviate class imbalance when dealing with symmetric noise in FL settings.

\begin{figure}[h!]
	\centering
        \includegraphics[width=1\columnwidth]{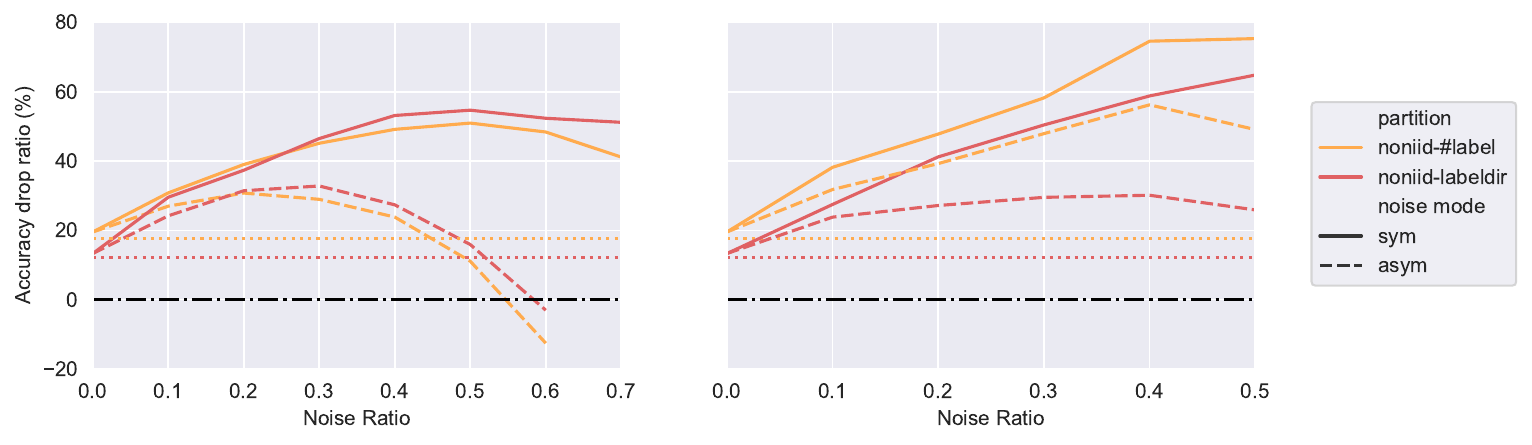}
	\caption{Accuracy drop ratio from IID to Non-IID with globalized noise on CIFAR-10. Give noise ratio $\varepsilon$, the accuracy drop ratio of certain \texttt{non-iid} partition at $\varepsilon$ is calculated by $[\mathrm{acc}_{\texttt{iid}}(\varepsilon) - \mathrm{acc}_{\texttt{non-iid}}(\varepsilon)] / \mathrm{acc}_{\texttt{iid}}(\varepsilon)$. The dotted lines mark the values when $\varepsilon=0.0$, and the dash-dotted lines mark accuracy drop ratio equals 0. $\varepsilon$ is $\varepsilon_{global}$ for globalized noise, and $\varepsilon_{local}$ for localized noise $\mathcal{U}(\varepsilon_{local}-0.1, \varepsilon_{local}+0.1)$. From left to right: globalized noise, localized noise.}
	\label{fig:cifar10-10-clients-global-local-iid-noniid-acc-drop-ratio}
\end{figure}

\textbf{2) Negative effect from label noise is not monotonic on Non-IID w.r.t. noise ratio.}
In FNLL where both Non-IID and noisy labels exist, how the noisy labels affect the model performance learning with Non-IID can be one of the most engaging problems. 
Here we denote $\mathrm{acc}_{\texttt{part}}(\varepsilon)$ as the test accuracy given noise with noise ratio $\varepsilon$ \footnote{$\varepsilon$ is $\varepsilon_{global}$ for globalized noise, and $\varepsilon_{local}$ for localized noise.} under \texttt{part} partition, and $\mathrm{acc}_{\texttt{part}}(0)$ for settings without label noise. 
We explore the accuracy drop ratio between Non-IID and IID with different noise ratios by calculating $[\mathrm{acc}_{\texttt{iid}}(\varepsilon) - \mathrm{acc}_{\texttt{non-iid}}(\varepsilon)] / \mathrm{acc}_{\texttt{iid}}(\varepsilon)$ for the chosen \texttt{non-iid} partition for noise ratio $\varepsilon$. The results are shown in Figure~\ref{fig:cifar10-10-clients-global-local-iid-noniid-acc-drop-ratio}. And the results for MNIST 10 clients are shown in Figure~\ref{fig:mnist-10-clients-global-local-iid-noniid-acc-drop-ratio} in Appendix~\ref{append:mnist-diff-noise}. 
Unsurprisingly, noisy labels aggravate the Non-IID problem as the noise ratio increases at first. 
However, the degradation effect stops enlarging when the noise ratio goes extremely large. 
For some settings, such degradation even decreases to become smaller than that with no label noise: the accuracy drop ratio of globalized asymmetric noise when $\varepsilon \ge 0.5$ is smaller than that with no label noise for two Non-IID cases.
Moreover, with $\varepsilon \ge 0.6$ in globalized asymmetric noise, noisy labels can even make Non-IID achieve higher performance than IID. 
This observation implies that label noise's negative effect on Non-IID situation is non-monotonicity w.r.t. noise ratio, which appeals to rethinking the role of label noise on Non-IID.

\textbf{3) Different sensitivity under different partitions.} We also explore the performance sensitivity of noise ratio in different FL partition settings. 
The performance sensitivity on the noise ratio of each partition is calculated using:
\begin{equation}
\label{eq:sensitivity}
    s(\varepsilon) = \left[\mathrm{acc}(\varepsilon) - \mathrm{acc}(\varepsilon + \Delta)\right] / \Delta,
\end{equation}
where $\mathrm{acc}(\cdot)$ is the test accuracy given noise ratio $\varepsilon$, and $\Delta > 0$ is the change of noise ratio. 
The larger $s(\varepsilon)$ is, the higher sensitivity the setting is to the current noise ratio $\varepsilon$. 
Note here a larger noise ratio always leads to worse performance in our setting, thus $\mathrm{acc}(\varepsilon) \ge \mathrm{acc}(\varepsilon + \Delta)$ holds in our experiment, that is, $s(\varepsilon) \ge 0$. 
Here we set $\Delta=0.1$ in Eq.~(\ref{eq:sensitivity}) for sensitivity calculation. 
The results  are shown in Figure~\ref{fig:sensitive-noise}. 
In both globalized and localized settings, Non-IID partitions become less sensitive to noise ratio when noise ratio increases, while on the contrary, IID becomes more sensitive as noise ratio grows for both symmetric and asymmetric noise. 
In other words, when the noise ratio is low, the performance in Non-IID settings will decrease significantly while decreases limitedly when the noise ratio is large. 
This indicates that when the noise ratio of an FL system is small, efforts on handling the label noise problem could lead to larger performance improvements; whereas it is better to focus on the Non-IID problem when confronted with a large amount of noisy labels.

% \begin{figure}[h]
%     \centering
%     \includegraphics[width=0.8\textwidth]{imgs/cifar10-10clients-global-local-sensitive-noise-ratio.pdf}
%     \caption{Sensitivity on different noise ratios on CIFAR-10 with 10 clients. The $x$-axis noise ratio is $\varepsilon_{global}$ for globalized noise, and $\varepsilon_{local}$ for localized noise $\mathcal{U}(\varepsilon_{local}-0.1, \varepsilon_{local}+0.1)$. From left to right: globalized noise, localized noise.}
%     \label{fig:sensitive-noise}
% \end{figure}

\textbf{4) Label noise increases the gradient norm in FL.} Previous CNLL works reveal that noisy labels can influence characteristics of the training process: \cite{han2018co, DBLP:conf/iclr/LiSH20} believe that noisy samples tend to have larger loss values compared with clean ones; \cite{pmlr-v139-liu21v, DBLP:journals/corr/abs-2212-04055} showed that noisy samples increase gradient norm of the training batch. 
Some FNLL papers have utilized sample loss for noisy sample detection in FNLL, while none has explored noisy labels' impact on gradient norm in FNLL yet. 
We calculate the per-round global model gradient norm as $\| w^t - w^{t-1} \|_2$ for each round and perform FedAvg training for \texttt{iid}, \texttt{noniid-\#label}, and \texttt{noniid-labeldir} with clean or globalized/localized noise. 
We use a learning rate of $0.005$ for all different settings for fair comparison in this experiment. 
Figure~\ref{fig:grad-norm-global} shows the results of globalized settings, and Figure~\ref{fig:grad-norm-local} in Appendix~\ref{append:grad-norm-local} shows the results of localized settings. 
We conclude from these two figures: 
\emph{1)} the gradient norm converges during the training process; 
\emph{2)} noisy labels enlarge the gradient norm of the global model under both IID and Non-IID partitions; 
\emph{3)} given the same noise setting, the curve of gradient norm under IID fluctuates significantly compared with Non-IID partitions. 
From these observations, we conjecture that the gradient norm is also ponderable information that can be leveraged in FNLL. 
\bibliographystyle{ACM-Reference-Format}
\bibliography{references}

%%
%% If your work has an appendix, this is the place to put it.
% \clearpage
\appendix

\section{Experiment details}
\label{append:experiment-detail}

We use VGG-16 \cite{DBLP:journals/corr/SimonyanZ14a} as the base model for CIFAR-10 and SVHN experiments, and use CNN for MNIST following \cite{DBLP:conf/icde/LiDCH22}. 
We use SGD optimizer with momentum $0.9$, default weight decay as $5\mathrm{e}{-4}$, and the learning rate is tuned from $\{0.01, 0.005, 0.002, 0.001\}$ for all settings. 
The batch size is set to 128 and the number of local epochs is set to 5 by default. 
For basic 20 settings (clean, $\varepsilon_{global}=0.4$ and $\mathcal{U}(0.3, 0.5)$ with 4 partitions) on all datasets, we use 500 rounds training.
For experiments with different noise ratios on CIFAR-10, we use 500 rounds for globalized noise, and 700 rounds for localized noise for convergence considerations. 

Our FNLL simulation pipelines are built upon FedLab \cite{JMLR:v24:22-0440} framework and PyTorch \cite{NEURIPS2019_9015}. And we release the codes with Apache License 2.0.

All the experiments are conducted on a cluster of 8 Tesla V100, 4 NVIDIA GTX 2080 Ti GPUs, and 4 NVIDIA GTX 1080 Ti GPUs.

\section{Data Partitions}
\label{append:data-partition}

Given a data sample $(x, y)$, non-IID indicates the joint probability distribution $\mathcal{P}(x, y)$ is different for different client $i$ and $j$ ($i, j \in [K]$, $i \neq j$), that is, $\mathcal{P}_i(x,y) \neq \mathcal{P}_j(x,y)$. By Bayes' theorem, $\mathcal{P}(x, y)$ can be rewritten as $\mathcal{P}(y|x) \mathcal{P}(x)$ and $\mathcal{P}(x|y) \mathcal{P}(y)$. Based on previous research \cite{DBLP:conf/aistats/McMahanMRHA17, DBLP:journals/ftml/KairouzMABBBBCC21, DBLP:conf/icde/LiDCH22}, we present IID and three representative non-IID schemes in \texttt{FedNoisy}. In this section, we first introduce IID partition, then partition methodologies for non-IID cases.

\textbf{IID} \;
In IID, data distributions on each client are independent and identical. Following \cite{DBLP:conf/aistats/McMahanMRHA17}, given a global dataset with $N$ training samples, we first randomly shuffle it, then partition it over $K$ clients, each receiving $\lfloor \frac{N}{K} \rfloor$ samples. Note that this partition keeps the original class ratio in the global dataset. If a class imbalance problem exists in the global dataset, then each client has the same class imbalance issue on its local dataset. %\footnote{WebVision is an imbalanced dataset.}. 
We use \texttt{iid} to denote the IID partition.

\textbf{Quantity skew} \;
In quantity distribution skew, different clients can hold different amounts of data, while keeping the same class distribution. For example, large academic hospitals usually have larger datasets compared with small community hospitals \cite{DBLP:journals/corr/abs-2107-08371}. Dirichlet distribution  \cite{DBLP:conf/icde/LiDCH22} and Log-Normal distribution \cite{DBLP:conf/iclr/AcarZNMWS21} can be used to simulate this Non-IID partition. Here we use Dirichlet distribution following \cite{DBLP:conf/icde/LiDCH22}. Given $K$ clients, we draw $K$-dim vector $\mathbf{q} \sim \mathrm{Dir}_K(\alpha)$ and allocate $\lfloor q_k N \rfloor$ samples from global dataset $\mathcal{D}$ to client $k$ for $k=1, \ldots,K$. The parameter $\alpha > 0$ can regulate the imbalance level of the quantity skew: a larger $\alpha$ will even lead to a partition result close to the IID partition. We use \texttt{noniid-quantity=$\alpha$} to denote quantity distribution skew with parameter value $\alpha$. 

\textbf{Label distribution skew} \;
In label distribution skew, the prior distributions $\mathcal{P}(y)$ may vary across clients, while $\mathcal{P}(x|y)$ is shared in the federated system. For example, primary angle closure glaucoma (PACG) is much more common than primary open-angle glaucoma (POAG) in the Asian population than African, Hispanic population \cite{Randhawabjophthalmol-2021-319470}. 
In \texttt{FedNoisy}, we include two common practices for label distribution skew, namely \emph{distribution-based label distribution skew} and \emph{quantity-based label distribution skew}. 
\begin{itemize}
    \item \emph{Dirichlet label skew} uses Dirichlet distribution to determine the class distribution on clients \cite{DBLP:conf/icml/YurochkinAGGHK19, DBLP:conf/iclr/WangYSPK20, DBLP:conf/icde/LiDCH22}. 
    Given $K$ clients and $C$ classes, we draw a $K$-dim vector $\mathbf{q}_i$ from $\mathrm{Dir}_K(\alpha)$ for each class $i \in [C]$ and allocate $q_{i k}$ proportion of global dataset with label $i$ to local client $k$. 
    In addition to various overall sample sizes, the clients also have varying class distributions.
    A larger $\alpha$ encourages the class distribution among clients close to a uniform distribution. We use \texttt{noniid-labeldir=$\alpha$} to denote distribution-based label distribution skew with parameter $\alpha$. 
    \item \emph{Label-quantity skew} assigns samples of specified number of classes on each client \cite{DBLP:conf/aistats/McMahanMRHA17, DBLP:conf/icde/LiDCH22, 10.1145/3286490.3286559}. 
    Here we follow the strategy in \cite{DBLP:conf/icde/LiDCH22}. Given the number of classes on each client as $c$, we first assign $c$ different classes to each client and then equally divide samples of each class according to the number of clients assigned to them. 
    When $c$ equals the total class number in the federation, the partition is equivalent to the IID partition. 
    We use \texttt{noniid-\#label=$c$} to denote quantity-based label distribution skew with parameter $c$.
\end{itemize}

\section{Fair comparison between globalized and localized noise scene}
\label{append:fair-comparison-local-global}
 
\begin{figure*}  %[h!]
    \centering
    \includegraphics[width=0.7\textwidth]{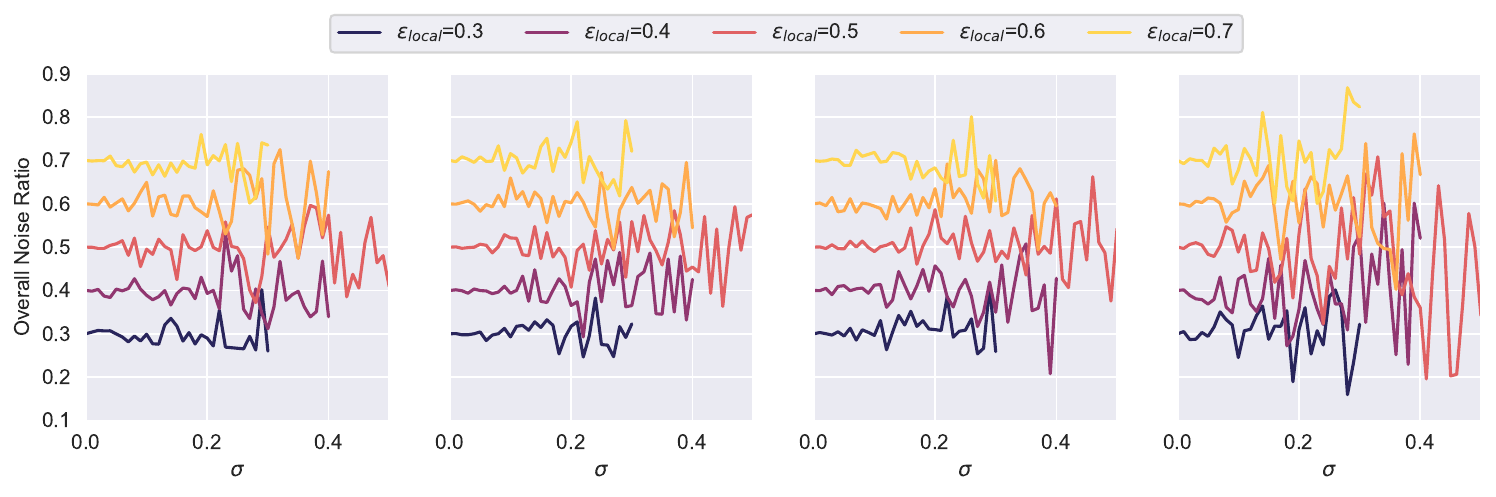}
    \caption{Overall noisy label ratio in localized symmetric noise $\mathcal{U}(\varepsilon_{local}-\sigma, \varepsilon_{local}+\sigma)$ on CIFAR-10 with 10 clients. From left to right: \texttt{iid}, \texttt{noniid-labeldir}, \texttt{noniid-\#label} and \texttt{noniid-labeldir} partition.}
    \label{fig:fednll-local-sym-cifar10}
\end{figure*}

Calculate the total noise ratio in the whole FL system in a localized noise setting. Local noise ratio is drawn from $\varepsilon_k \sim \mathcal{U}(\varepsilon_{min}, \varepsilon_{max})$. Here we let $\varepsilon_{min} = \varepsilon_{local} - \sigma$, and $\varepsilon_{max} = \varepsilon_{local} + \sigma$, where $\varepsilon_{local}$ is the mean value for the uniform distribution $\mathcal{U}(\varepsilon_{min}, \varepsilon_{max})$. Check the relation between the total FL noise ratio and $\varepsilon_{local}$. Figure~\ref{fig:fednll-local-sym-cifar10} indicates that, when $\sigma$ is small (less than $0.1$), the total noise sample ratio in localized noise setting, where local noise ratio follows distribution $\mathcal{U}(\varepsilon_{local} - \sigma, \varepsilon_{local} + \sigma)$, is always close to the mean local noise ratio for all clients, that is, $\varepsilon_{local}$. When $\sigma$ is large, the total noise ratio is more likely to deviate from $\varepsilon_{local}$. Motivated by this, we will compare the results between localized noise and globalized noise when $\sigma \le 0.1$, while dismissing such comparison when $\sigma > 0.1$.

\section{Gradient norm under localized noise setting}
\label{append:grad-norm-local}

\begin{figure*}  %[h]
    \centering
    \includegraphics[width=\textwidth]{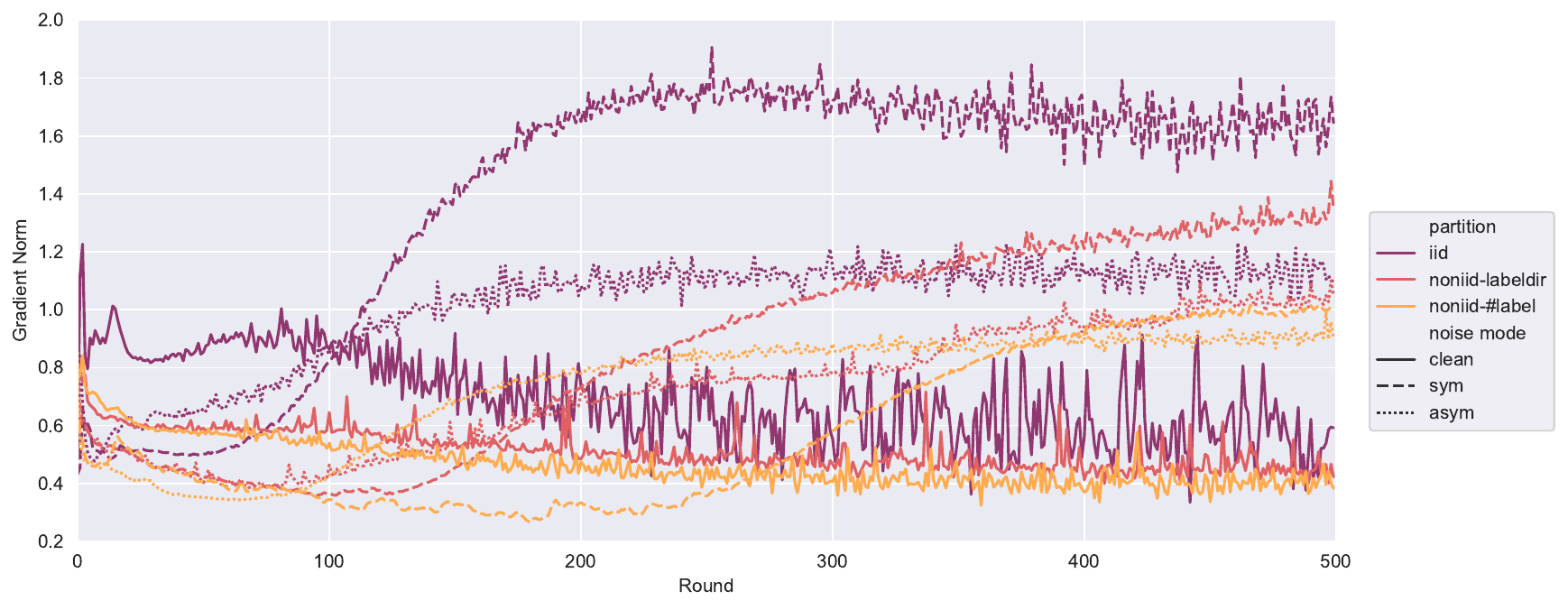}
    \caption{Gradient norm of global model for clean and localized noise $\varepsilon_k \sim \mathcal{U}(0.3, 0.5)$.}
    \label{fig:grad-norm-local}
\end{figure*}

CIFAR-10 10 clients with localized label noise, using FedAvg. The gradient norm result is shown in Figure~\ref{fig:grad-norm-local}.

\section{Results on other datasets}
\label{append:other-dataset-results}

\subsection{MNIST results on FedAvg}
\label{append:mnist-results}

FedAvg results on MNIST, 500 rounds, 5 local epochs, SGD, 10 clients with $100\%$ selection ratio. The partition is \texttt{iid}, \texttt{noniid-labeldir=0.1}, \texttt{noniid-\#label=3} and \texttt{noniid-quantity=0.1}. Results are shown in Table~\ref{tab:mnist-10-clients-fedavg}.

\begin{table*}  %[h]
\centering
\caption{Test accuracy of MNIST on 10 clients using FedAvg with CNN}
\label{tab:mnist-10-clients-fedavg}
\scalebox{0.78}{
\begin{tabular}{ccccccc} 
\toprule
\multirow{2}{*}{}           & \multirow{2}{*}{\textbf{Noise mode}} & \multirow{2}{*}{\textbf{Noise ratio}}    & \multicolumn{4}{c}{\textbf{Data partition}}                                    \\ 
\cmidrule(lr){4-4}\cmidrule(lr){5-5}\cmidrule(lr){6-6}\cmidrule(lr){7-7}
                            &                                      &                                          & \texttt{iid}      & \texttt{noniid-\#label}    & \texttt{noniid-labeldir}   & \texttt{noniid-quantity}    \\ 
\midrule
                            & Clean                                & $0.0$                                    & $99.17 \pm 0.03$  & $98.49 \pm 0.29$            & $98.05 \pm 0.53$            & $99.16 \pm 0.04$   \\ 
\midrule
\multirow{2}{*}{Globalized} & Sym.                                 & \multirow{2}{*}{$0.4$}                   & $71.39 \pm 0.52$  & $39.22 \pm 1.48$            & $40.70 \pm 6.47$            & $71.31 \pm 1.92$  \\ 
\cmidrule(lr){2-2}
                            & Asym.                                &                                          & $61.88 \pm 1.04$  & $53.02 \pm 1.05$            & $54.20 \pm 3.23$            & $60.61 \pm 0.63$  \\ 
\midrule
\multirow{2}{*}{Localized}  & Sym.                                 & \multirow{2}{*}{$\mathcal{U}(0.3, 0.5)$} & $70.88 \pm 0.55$  & $41.98 \pm 1.25$            & $40.87 \pm 4.32$            & $69.56 \pm 5.64$  \\ 
\cmidrule(lr){2-2}
                            & Asym.                                &                                          & $61.28 \pm 1.75$  & $59.78 \pm 2.19$            & $61.46 \pm 4.60$            & $57.84 \pm 8.80$  \\
\bottomrule
\end{tabular}}
\end{table*}

\subsection{SVHN results}

FedAvg results on SVHN, 500 rounds, 5 local epochs, SGD, 10 clients with $100\%$ selection ratio. The partition setting is \texttt{iid}, \texttt{noniid-\#label=3}, \texttt{noniid-labeldir=0.1}, \texttt{noniid-quantity=0.1}. Results are shown in Table~\ref{tab:svhn-10-clients-fedavg}.

\begin{table*}%[h]
\centering
\caption{Test accuracy of SVHN on 10 clients using FedAvg with VGG16}
\label{tab:svhn-10-clients-fedavg}
\scalebox{0.78}{
\begin{tabular}{ccccccc} 
\toprule
\multirow{2}{*}{}           & \multirow{2}{*}{\textbf{Noise mode}} & \multirow{2}{*}{\textbf{Noise ratio}}    & \multicolumn{4}{c}{\textbf{Data partition}}                                    \\ 
\cmidrule(lr){4-4}\cmidrule(lr){5-5}\cmidrule(lr){6-6}\cmidrule(lr){7-7}
                            &                                      &                                          & \texttt{iid}      & \texttt{noniid-\#label}    & \texttt{noniid-labeldir}   & \texttt{noniid-quantity}    \\ 
\midrule
                            & Clean                                & $0.0$                                    & $95.92$  & $40.80$            & $47.37$            & $95.54$   \\ 
\midrule
\multirow{2}{*}{Globalized} & Sym.                                 & \multirow{2}{*}{$0.4$}                   & $78.32$  & $19.59$            & $19.59$            & $72.96$  \\ 
\cmidrule(lr){2-2}
                            & Asym.                                &                                          & $62.79$  & $38.85$            & $43.23$            & $61.28$  \\ 
\midrule
\multirow{2}{*}{Localized}  & Sym.                                 & \multirow{2}{*}{$\mathcal{U}(0.3, 0.5)$} & $76.99$  & $10.90$            & $8.22$            & $72.33$  \\ 
\cmidrule(lr){2-2}
                            & Asym.                                &                                          & $60.76$  & $22.13$            & $41.47$            & $60.67$  \\
\bottomrule
\end{tabular}}
\end{table*}

\subsection{CIFAR-100 results}

FedAvg results on CIFAR-100, 500 rounds, 5 local epochs, SGD, 10 clients with $100\%$ selection ratio. Results are shown in Table~\ref{tab:cifar100-10-clients-fedavg}. The partition setting is \texttt{iid}, \texttt{noniid-\#label=50}, \texttt{noniid-labeldir=0.1}, \texttt{noniid-quantity=0.1}.

\begin{table*} %[h]
\centering
\caption{Test accuracy of CIFAR-100 on 10 clients using FedAvg with ResNet-34}
\label{tab:cifar100-10-clients-fedavg}
\scalebox{0.78}{
\begin{tabular}{ccccccc} 
\toprule
\multirow{2}{*}{}           & \multirow{2}{*}{\textbf{Noise mode}} & \multirow{2}{*}{\textbf{Noise ratio}}    & \multicolumn{4}{c}{\textbf{Data partition}}                                    \\ 
\cmidrule(lr){4-4}\cmidrule(lr){5-5}\cmidrule(lr){6-6}\cmidrule(lr){7-7}
                            &                                      &                                          & \texttt{iid}      & \texttt{noniid-\#label}    & \texttt{noniid-labeldir}   & \texttt{noniid-quantity}    \\ 
\midrule
                            & Clean                                & $0.0$                                    & $74.15 \pm 0.11$  & $71.58 \pm 0.31$ & $68.31 \pm 1.97$ & $72.08 \pm 1.80$   \\ 
\midrule
\multirow{2}{*}{Globalized} & Sym.                                 & \multirow{2}{*}{$0.4$}                   & $43.81 \pm 0.74$  & $37.99 \pm 1.12$            & $29.2 \pm 0.64$            & $42.51 \pm 2.68$ \\ 
\cmidrule(lr){2-2}
                            & Asym.                                &                                          & $44.50 \pm 0.60$  & $40.48 \pm 0.44$            & $36.35 \pm 0.48$            & $41.47 \pm 1.56$  \\ 
\midrule
\multirow{2}{*}{Localized}  & Sym.                                 & \multirow{2}{*}{$\mathcal{U}(0.3, 0.5)$} & $44.53 \pm 0.9$  & $36.28 \pm 3.51$           & $28.52 \pm 1.32$           & $42.01 \pm 3.32$  \\ 
\cmidrule(lr){2-2}
                            & Asym.                                &                                          & $44.96 \pm 0.63$  & $47.00 \pm 3.36$      & $40.10 \pm 1.72$            & $40.13 \pm 3.52$  \\
\bottomrule
\end{tabular}}
\end{table*}

\subsection{Clothing1M results}

FedAvg results on Clothing1M using ResNet-50 \cite{DBLP:conf/cvpr/HeZRS16}, 150 rounds, 5 local epochs, SGD, 10 clients with $100\%$ selection ratio. The partition setting is \texttt{iid}, \texttt{noniid-\#label=5}, \texttt{noniid-labeldir=0.1}, \texttt{noniid-quantity=0.1}. Results are shown in Table~\ref{tab:clothing1m-10-clients-fedavg}.

\begin{table*}  %[h]
\centering
\caption{Test accuracy of Clothing1M on 10 clients using FedAvg with ResNet-50}
\label{tab:clothing1m-10-clients-fedavg}
\scalebox{0.78}{
\begin{tabular}{cccccc} 
\toprule
\multirow{2}{*}{}           & \multirow{2}{*}{\textbf{Noise mode}} &  \multicolumn{4}{c}{\textbf{Data partition}}                                    \\ 
\cmidrule(lr){3-3}\cmidrule(lr){4-4}\cmidrule(lr){5-5}\cmidrule(lr){6-6}                                &                                          & \texttt{iid}      & \texttt{noniid-\#label}    & \texttt{noniid-labeldir}   & \texttt{noniid-quantity}    \\ 
\midrule
                            & Real                                    & $70.99 \pm 0.33$  & $53.68 \pm 5.26$ & $48.41 \pm 9.52$ & $67.70 \pm 0.87$   \\
\bottomrule
\end{tabular}}
\end{table*}

\section{Different noise ratio experiment results}

\subsection{CIFAR-10 results}
\label{append:cifar10-diff-noise}

The partitions are \texttt{iid}, \texttt{noniid-\#label=3} and \texttt{noniid-labeldir=0.1} partitions. Table~\ref{tab:cifar10-global-diff-noise} presents results under globalized noise with different noise ratios for 500 rounds of FedAvg training on 10 clients. Table~\ref{tab:cifar10-local-diff-noise} presents results under localized noise with different noise ratios for 700 rounds of FedAvg training on 10 clients.

\begin{table*} %[h]
\centering
\caption{Test accuracy for globalized noise with different noise ratios on CIFAR-10 using FedAvg}
\label{tab:cifar10-global-diff-noise}
\scalebox{0.8}{
\begin{tabular}{ccccc} 
\toprule
\multirow{2}{*}{\textbf{Noise mode}} & \multirow{2}{*}{\textbf{Noise ratio}} & \multicolumn{3}{c}{\textbf{Data partition}}                                                                                                                                                                                                                   \\ 
\cmidrule(lr){3-3}\cmidrule(lr){4-4}\cmidrule(lr){5-5}
                                     &                                       & \texttt{iid}            & \texttt{noniid-\#label}  & \texttt{noniid-labeldir}                                                                 \\ 
\midrule
Clean                                & 0.0                                   & $90.82 \pm 0.14$ & $73.09 \pm 4.33$    & $78.73 \pm 3.56$  \\ 
\midrule
\multirow{7}{*}{Sym.}                & 0.1                                   & $85.86 \pm 0.16$ & $59.40 \pm 1.80$    & $60.48 \pm 3.57$  \\ 
\cmidrule{2-2}
                                     & 0.2                                   & $80.73 \pm 0.24$ & $49.23 \pm 1.43$    & $50.58 \pm 7.08$  \\ 
\cmidrule{2-2}
                                     & 0.3                                   & $73.92 \pm 0.21$ & $40.56 \pm 1.23$    & $39.55 \pm 5.74$  \\ 
\cmidrule{2-2}
                                     & 0.4                                   & $64.96 \pm 0.39$ & $33.04 \pm 1.86$    & $30.44 \pm 6.30$  \\ 
\cmidrule{2-2}
                                     & 0.5                                   & $53.54 \pm 1.30$ & $26.26 \pm 1.69$    & $24.26 \pm 4.20$  \\ 
\cmidrule{2-2}
                                     & 0.6                                   & $39.88 \pm 0.74$ & $20.59 \pm 0.87$    & $19.00 \pm 2.44$  \\ 
\cmidrule{2-2}
                                     & 0.7                                   & $26.31 \pm 1.27$ & $15.48 \pm 0.13$    & $12.84 \pm 1.98$  \\ 
\midrule
\multirow{6}{*}{Asym.}               & 0.1                                   & $86.18 \pm 0.23$ & $62.93 \pm 4.63$    & $65.33 \pm 4.64$  \\ 
\cmidrule{2-2}
                                     & 0.2                                   & $80.01 \pm 0.39$ & $55.39 \pm 4.27$    & $54.86 \pm 5.36$  \\ 
\cmidrule{2-2}
                                     & 0.3                                   & $70.23 \pm 0.47$ & $49.90 \pm 3.25$    & $47.19 \pm 6.61$  \\ 
\cmidrule{2-2}
                                     & 0.4                                   & $57.84 \pm 0.59$ & $44.06 \pm 3.65$    & $42.02 \pm 5.51$  \\ 
\cmidrule{2-2}
                                     & 0.5                                   & $44.34 \pm 0.24$ & $39.42 \pm 2.61$    & $37.28 \pm 3.52$  \\ 
\cmidrule{2-2}
                                     & 0.6                                   & $31.06 \pm 0.10$ & $34.96 \pm 2.21$    & $32.00 \pm 3.36$  \\
\bottomrule
\end{tabular}}
\end{table*}

\begin{table*}  %[h!]
\centering
\caption{Test accuracy for localized noise with different noise ratios on CIFAR-10 using FedAvg}
\label{tab:cifar10-local-diff-noise}
\scalebox{0.8}{
\begin{tabular}{ccccc} 
\toprule
\multirow{2}{*}{\textbf{Noise mode}} & \multirow{2}{*}{\textbf{Noise ratio}} & \multicolumn{3}{c}{\textbf{Data partition}}              \\ 
\cmidrule(lr){3-3}\cmidrule(lr){4-4}\cmidrule(lr){5-5}
                                     &                                       & \texttt{iid}              & \texttt{noniid-\#label}   & \texttt{noniid-labeldir}    \\ 
\midrule
\multirow{5}{*}{Sym.}                & $\mathcal{U}(0.0, 0.2)$                                   & $86.26 \pm 0.08$ & $53.31 \pm 4.07$ & $62.56 \pm 11.59$  \\ 
\cmidrule{2-2}
                                     & $\mathcal{U}(0.1, 0.3)$                                   & $80.91 \pm 0.51$ & $42.24 \pm 1.59$ & $47.57 \pm 11.96$  \\ 
\cmidrule{2-2}
                                     & $\mathcal{U}(0.2, 0.4)$                                   & $74.27 \pm 0.70$ & $31.02 \pm 4.32$ & $36.82 \pm 10.65$  \\ 
\cmidrule{2-2}
                                     & $\mathcal{U}(0.3, 0.5)$                                   & $66.05 \pm 0.49$ & $16.78 \pm 1.58$ & $27.21 \pm 9.08$   \\ 
\cmidrule{2-2}
                                     & $\mathcal{U}(0.4, 0.6)$                                   & $54.31 \pm 1.96$ & $13.39 \pm 3.14$ & $19.10 \pm 7.17$   \\ 
\midrule
\multirow{5}{*}{Asym.}               & $\mathcal{U}(0.0, 0.2)$                                   & $86.99 \pm 0.18$ & $59.34 \pm 4.69$ & $66.27 \pm 9.20$   \\ 
\cmidrule{2-2}
                                     & $\mathcal{U}(0.1, 0.3)$                                   & $80.32 \pm 0.76$ & $48.78 \pm 1.81$ & $58.49 \pm 5.70$   \\ 
\cmidrule{2-2}
                                     & $\mathcal{U}(0.2, 0.4)$                                   & $69.53 \pm 0.39$ & $36.21 \pm 3.74$ & $49.00 \pm 8.07$   \\ 
\cmidrule{2-2}
                                     & $\mathcal{U}(0.3, 0.5)$                                   & $56.96 \pm 2.77$ & $24.93 \pm 3.18$ & $39.80 \pm 6.86$   \\ 
\cmidrule{2-2}
                                     & $\mathcal{U}(0.4, 0.6)$                                   & $45.34 \pm 1.06$ & $23.09 \pm 1.85$ & $33.60 \pm 6.62$   \\
\bottomrule
\end{tabular}}
\end{table*}

\begin{table*} %[h]
\centering
\caption{Test accuracy for globalized noise with different noise ratios on MNIST using FedAvg}
\label{tab:mnist-global-diff-noise}
\scalebox{0.8}{
\begin{tabular}{ccccc} 
\toprule
\multirow{2}{*}{\textbf{Noise mode}} & \multirow{2}{*}{\textbf{Noise ratio}} & \multicolumn{3}{c}{\textbf{Data partition}}                                                                                                                                                                                                                   \\ 
\cmidrule(lr){3-3}\cmidrule(lr){4-4}\cmidrule(lr){5-5}
                                     &                                       & \texttt{iid}            & \texttt{noniid-\#label}  & \texttt{noniid-labeldir}                                                                 \\ 
\midrule
Clean                                & 0.0                                   & $99.17  \pm  0.03$ & $98.49  \pm  0.29$   & $98.05  \pm  0.53$  \\ 
\midrule
\multirow{7}{*}{Sym.}                & 0.1                                   & $95.67 \pm 0.07$ & $74.95 \pm 1.03$    & $75.4 \pm 5.73$  \\ 
\cmidrule{2-2}
                                     & 0.2                                   & $89.63 \pm 0.13$ & $57.84 \pm 0.93$    & $58.88 \pm 5.83$  \\ 
\cmidrule{2-2}
                                     & 0.3                                   & $81.2 \pm 0.02$ & $46.72 \pm 1.55$    & $49.09 \pm 6.4$  \\ 
\cmidrule{2-2}
                                     & 0.4                                   & $71.75 \pm 0.67$ & $39.02 \pm 1.76$    & $40.49 \pm 5.57$  \\ 
\cmidrule{2-2}
                                     & 0.5                                   & $60.03 \pm 0.6$ & $33.57 \pm 1.45$    & $34.62 \pm 4.48$  \\ 
\cmidrule{2-2}
                                     & 0.6                                   & $48.97 \pm 1.74$ & $29.5 \pm 0.84$   & $27.68 \pm 2.78$ \\ 
\cmidrule{2-2}
                                     & 0.7                                   & $37.33 \pm 0.9$ & $22.37 \pm 0.88$   & $22.86 \pm 2.78$ \\ 
\midrule
\multirow{6}{*}{Asym.}               & 0.1                                   & $94.78 \pm 0.16$ & $82.37 \pm 0.21$    & $93.66 \pm 0.44$ \\ 
\cmidrule{2-2}
                                     & 0.2                                   & $85.38 \pm 0.03$ & $70.45 \pm 0.19$    & $83.66 \pm 0.72$  \\ 
\cmidrule{2-2}
                                     & 0.3                                   & $74.55 \pm 0.76$ & $61.9 \pm 1.66$    & $72.16 \pm 0.45$  \\ 
\cmidrule{2-2}
                                     & 0.4                                   & $61.59 \pm 0.55$ & $53.14 \pm 1.24$    & $60.97 \pm 13.11$  \\ 
\cmidrule{2-2}
                                     & 0.5                                   & $48.31 \pm 0.23$ & $46.98 \pm 1.23$    & $48.4 \pm 0.58$  \\ 
\cmidrule{2-2}
                                     & 0.6                                   & $35.18 \pm 0.66$ & $37.88 \pm 0.84$    & $36.38 \pm 1.12$ \\
\bottomrule
\end{tabular}}
\end{table*}

\begin{table*} %[h!]
\centering
\caption{Test accuracy for localized noise with different noise ratios on MNIST using FedAvg}
\label{tab:mnist-local-diff-noise}
\scalebox{0.8}{
\begin{tabular}{ccccc} 
\toprule
\multirow{2}{*}{\textbf{Noise mode}} & \multirow{2}{*}{\textbf{Noise ratio}} & \multicolumn{3}{c}{\textbf{Data partition}}              \\ 
\cmidrule(lr){3-3}\cmidrule(lr){4-4}\cmidrule(lr){5-5}
                                     &                                       & \texttt{iid}              & \texttt{noniid-\#label}   & \texttt{noniid-labeldir}    \\ 
\midrule
\multirow{5}{*}{Sym.}                & $\mathcal{U}(0.0, 0.2)$                                   & $95.26 \pm 0.54$ & $82.79 \pm 1.95$ & $80.89 \pm 3.16$  \\ 
\cmidrule{2-2}
                                     & $\mathcal{U}(0.1, 0.3)$                                   & $89.36 \pm 0.73$ & $67.62 \pm 2.32$ & $62.81 \pm 4.95$ \\ 
\cmidrule{2-2}
                                     & $\mathcal{U}(0.2, 0.4)$                                   & $80.96 \pm 1.14$ & $54.45 \pm 1.75$ & $50.75 \pm 4.91$  \\ 
\cmidrule{2-2}
                                     & $\mathcal{U}(0.3, 0.5)$                                   & $70.99 \pm 0.43$ & $42.73 \pm 0.96$ & $40.64 \pm 3.93$   \\ 
\cmidrule{2-2}
                                     & $\mathcal{U}(0.4, 0.6)$                                   & $59.69 \pm 1.42$ & $32.19 \pm 1.23$ & $32.21 \pm 3.91$   \\ 
\midrule
\multirow{5}{*}{Asym.}               & $\mathcal{U}(0.0, 0.2)$                                   & $94.64 \pm 1.13$ & $86.91 \pm 1.18$ & $87.30 \pm 2.97$   \\ 
\cmidrule{2-2}
                                     & $\mathcal{U}(0.1, 0.3)$                                   & $85.36 \pm 1.57$ & $76.67 \pm 0.78$ & $76.11 \pm 4.63$  \\ 
\cmidrule{2-2}
                                     & $\mathcal{U}(0.2, 0.4)$                                   & $73.75 \pm 0.97$ & $66.97 \pm 0.94$ & $68.57 \pm 4.61$   \\ 
\cmidrule{2-2}
                                     & $\mathcal{U}(0.3, 0.5)$                                   & $61.11 \pm 1.87$ & $59.70 \pm 2.32$ & $61.06 \pm 4.21$   \\ 
\cmidrule{2-2}
                                     & $\mathcal{U}(0.4, 0.6)$                                   & $47.95 \pm 1.05$ & $53.61 \pm 2.79$ & $54.71 \pm 4.84$   \\
\bottomrule
\end{tabular}}
\end{table*}

\subsection{MNIST results}
\label{append:mnist-diff-noise}

\begin{figure*} %[h!]
	\centering
        \includegraphics[width=0.75\textwidth]{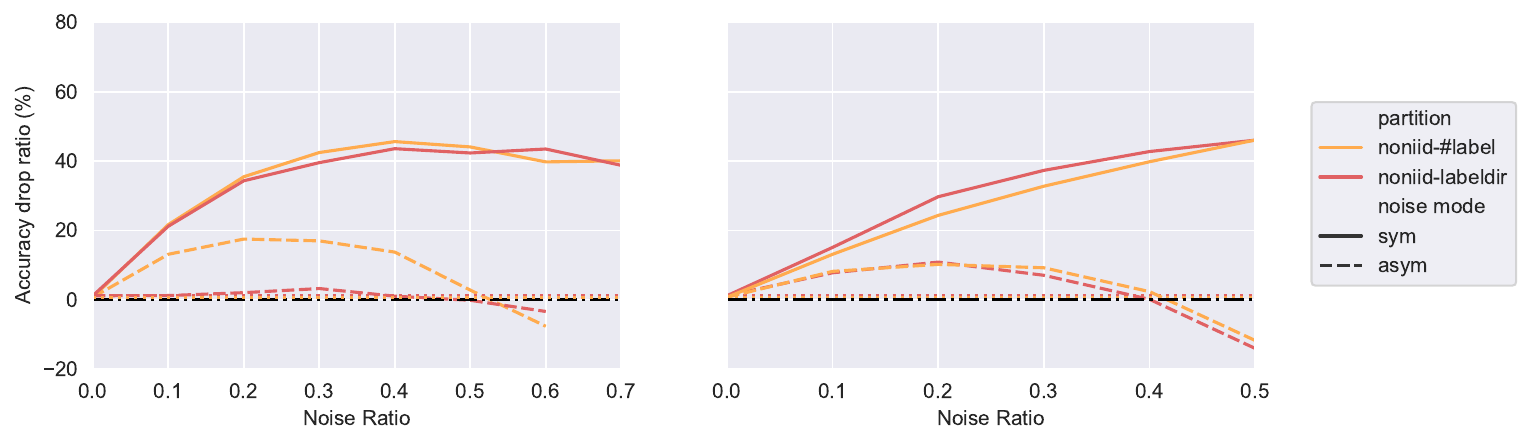}
	\caption{Accuracy drop ratio from IID to Non-IID with globalized noise on MNIST. Give noise ratio $\varepsilon$, the accuracy drop ratio of certain \texttt{non-iid} partition at $\varepsilon$ is calculated by $[\mathrm{acc}_{\texttt{iid}}(\varepsilon) - \mathrm{acc}_{\texttt{non-iid}}(\varepsilon)] / \mathrm{acc}_{\texttt{iid}}(\varepsilon)$. The dotted lines mark the values when $\varepsilon=0.0$, and the dash-dotted lines mark accuracy drop ratio equals 0. $\varepsilon$ is $\varepsilon_{global}$ for globalized noise, and $\varepsilon_{local}$ for localized noise $\mathcal{U}(\varepsilon_{local}-0.1, \varepsilon_{local}+0.1)$. From left to right: globalized noise, localized noise.}
	\label{fig:mnist-10-clients-global-local-iid-noniid-acc-drop-ratio}
\end{figure*}

The partitions are \texttt{iid}, \texttt{noniid-\#label=3} and \texttt{noniid-labeldir=0.1} partitions. Table~\ref{tab:mnist-global-diff-noise} presents results under globalized noise with different noise ratios for 500 rounds of FedAvg training on 10 clients. Table~\ref{tab:mnist-local-diff-noise} presents results under localized noise with different noise ratios for 500 rounds of FedAvg training on 10 clients. Figure~\ref{fig:mnist-10-clients-global-local-iid-noniid-acc-drop-ratio} presents the plot of the accuracy drop ratio between Non-IID and IID with different noise ratios.

\section{More Baseline Results}
\label{rebut:baseline}

\begin{figure*}  %[h!]
    \centering
    \includegraphics[width=0.7\textwidth]{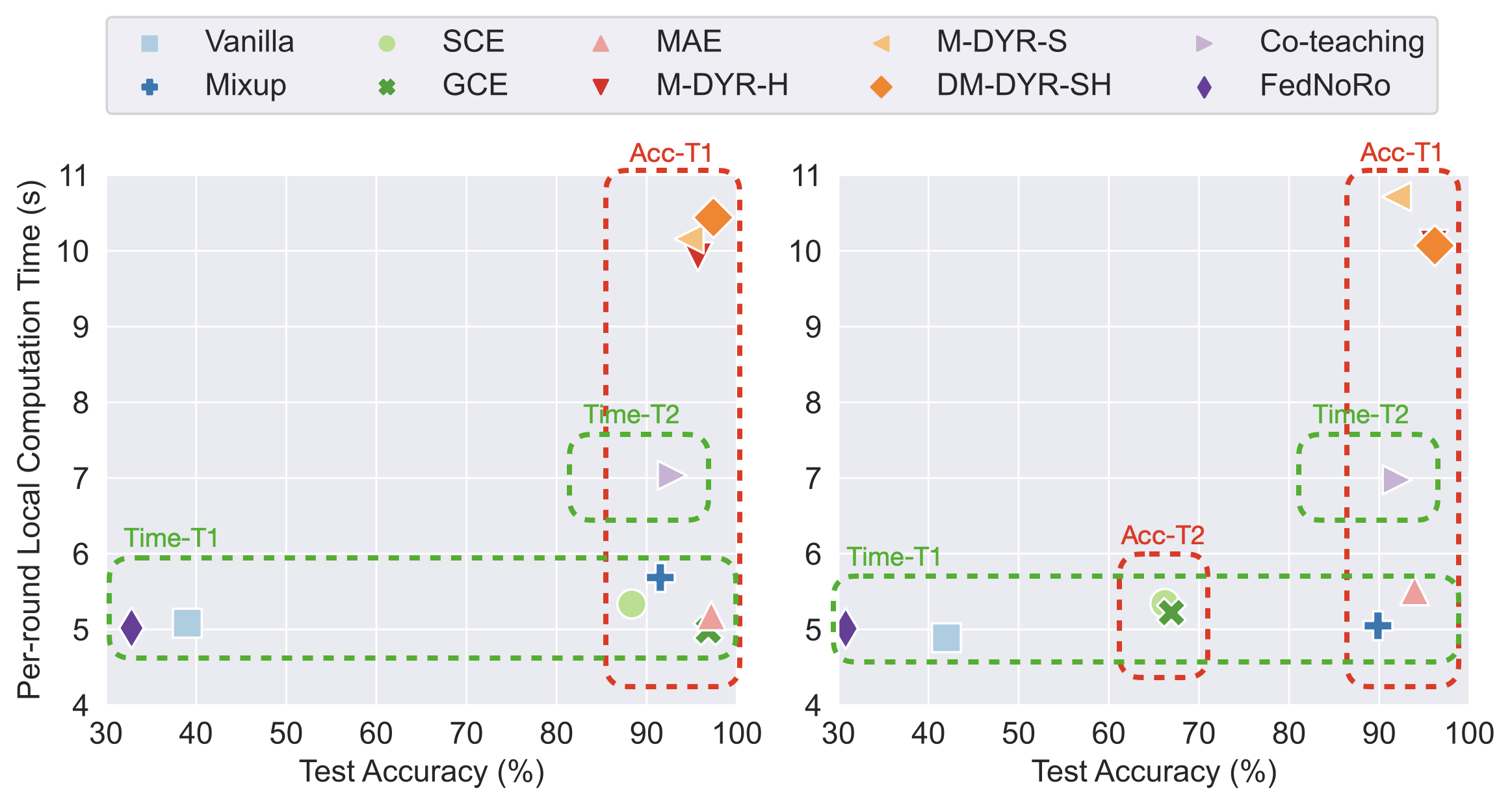}
    \caption{The trade-off between accuracy and per-round local computation time on MNIST on 10 clients under \texttt{noniid-\#label=3} partition with symmetric noise. From left to right: globalized noise, localized noise. 
    % Red and green dash boxes are used to categorize the baselines into several tiers for both test accuracy and per-round time, respectively. \textit{T1} denotes tier 1, and \textit{T2} denotes tier 2.
    }
    \label{fig:baseline-results-mnist}
\end{figure*}

The visualization of accuracy and computation trade-off on MNIST is shown in~\autoref{fig:baseline-results-mnist}.

\end{document}